\pdfoutput=1

\documentclass[11pt]{article}

\usepackage[]{acl}

\usepackage{times}
\usepackage{latexsym}
\usepackage{multirow}
\usepackage{booktabs}
\usepackage{tabularx}
\usepackage{makecell}
\usepackage{graphicx}
\usepackage{float}
\usepackage{algorithm}
\usepackage{hyperref}
\usepackage{xcolor}		
 \definecolor{darkblue}{rgb}{0, 0, 0.5}
 \hypersetup{colorlinks=true,citecolor=darkblue, linkcolor=darkblue, urlcolor=darkblue}
 
\def\Snospace~{\S{}} 

\usepackage{url}
\usepackage{graphicx}
\usepackage{bbding}
\usepackage[noend]{algpseudocode}
\usepackage{mathtools, nccmath}
\usepackage{colortbl}
\usepackage[normalem]{ulem}

\def\EDdel#1{\bgroup\markoverwith{\textcolor{cyan}{\rule[0.5ex]{2pt}{1pt}}}\ULon{#1}}

\def\daviddel#1{\bgroup\markoverwith{\textcolor{purple}{\rule[0.5ex]{2pt}{1pt}}}\ULon{#1}}

\usepackage[T1]{fontenc}

\usepackage[utf8]{inputenc}

\usepackage{microtype}
\usepackage{subfig}
\usepackage{amsmath}

\title{
Improving Chess Commentaries by Combining Language Models with Symbolic Reasoning Engines
}

\author{Andrew Lee$^{1,2}$ \hspace*{.3em} David Wu$^2$ \hspace*{.3em} Emily Dinan$^2$ \hspace*{.3em} Mike Lewis$^2$ \\
    ${}^1$ University of Michigan \hspace{.18in}
    ${}^2$ Meta AI \\
    \texttt{ajyl@umich.edu} \hspace*{.3em} \texttt{\{dwu,edinan,mikelewis\}@meta.com} \\}

\begin{document}

\maketitle

\begin{abstract}
\fontdimen2\font=2pt

Despite many recent advancements in language modeling, state-of-the-art language models lack grounding in the real world and struggle with tasks involving complex reasoning.
Meanwhile, advances in the symbolic reasoning capabilities of AI have led to systems that outperform humans in games like chess and Go \citep{alphazero}.
Chess commentary provides an interesting domain for bridging these two fields of research, as it requires reasoning over a complex board state and providing analyses in natural language.
In this work we demonstrate how to combine symbolic reasoning engines with controllable language models to generate chess commentaries.
We conduct experiments to demonstrate that our approach generates commentaries that are preferred by human judges over previous baselines.

\end{abstract}


\section{Introduction}
\label{sec:intro}

While large-scale language models continue to achieve impressive feats \citep{brown2020language, zhang2022opt}, they often generate plausible responses that share no relationship with the true state of the world \citep{bender-koller-2020-climbing, tacl-begin} and struggle with logical and symbolic reasoning tasks \citep{rae2021scaling, wei2022chain}.

Meanwhile, researchers have demonstrated symbolic reasoning systems that play games such as chess or Go at a higher level than humans or purely neural models \citep{alphazero, mcilroyyoung2020maia}, with evidence that they learn interpretable strategic concepts during training \citep{mcgrath2021acquisition}.
In this work, we investigate how to leverage the strengths of language models and these symbolic reasoning engines to generate accurate and strategically sound chess commentaries \citep{jhamtani-etal-2018-learning}.

The goal of chess commentary generation is to provide accurate descriptions or in-depth analyses of moves being played in a game.
Such a task provides an interesting domain for bridging the two research areas, as it requires a language model to understand the symbolic structure, rules, and strategies of chess, while correctly interpreting and predicting current and future game-states.

In this work, we combine symbolic reasoning engines with a controllable language model to generate chess commentaries.
Unlike prior work that either builds custom features \citep{jhamtani-etal-2018-learning} or custom neural architectures \citep{zang-etal-2019-automated}, we simply control our language model's behavior using signals produced by chess engines.

Concretely, during training, given a game-state and its corresponding commentary, we extract ``control codes'', or ``tags'', that correspond to information contained in the comment, such as quality of a move or the best line of moves.
Our commentary generation model learns to predict text conditioned on the extracted control codes.
These control codes provide an interface to integrate with a chess engine.
During inference, given a new game-state, we use the engine to supply control codes to the commentary generation model to control its output.
This approach not only relieves the language model of any reasoning tasks, but also allows our commentary generation model to incorporate super-human level insights, even if its training data consisted of commentary from weaker players.

We conduct multiple experiments, using both automated metrics and human evaluations to measure the performance of our approach over a few baselines.
We demonstrate that our approach generates preferable commentaries according to human judges.

Lastly, we provide detailed qualitative and error analyses and discuss open challenges.


\section{Related work}
\label{sec:related_work}

In this section we discuss two streams of related work: chess commentary generation, followed by relevant approaches for language models. 

\subsection{Chess commentary generation}
\label{subsec:related_work_chess_comm_gen}

Chess commentary generation was first proposed by \citet{jhamtani-etal-2018-learning}.
The authors craft features to represent chess game-states, which is attended to by a decoder model.
\citet{zang-etal-2019-automated} extend this work by using AlphaZero \citep{alphazero} to encode the game-state.
Similar tasks include Shogi commentary generation \citep{hirotaka}.
While prior work use hand-crafted features or neural architectures, we control our language model using the output of chess engines.

\subsection{Controlled text generation}
\label{subsec:related_work_controlled_text_gen}
Controlled text generation often involves conditioning a language model with specific signals, often special tokens, in order to guide the model's behavior.
By doing so, researchers demonstrate control over various aspects, such as style \citep{smith2020controlling}, content \citep{gupta-etal-2021-controlling, keskar2019ctrl}, or task-specific behavior \citep{radford2019language, hosseini, peng-etal-2021-soloist}.

\subsection{Reasoning systems with language models}
\label{subsec:related_work_reasoning_sys}
Researchers have successfully combined symbolic reasoning agents with language generation models for different tasks.
\citet{doi:10.1126/science.ade9097} builds a language model that interfaces with a strategic reasoning model to play Diplomacy, a dialogue-based board game. 
\citet{he-etal-2018-decoupling} and \citet{pmlr-v80-yarats18a} build competitive negotiating agents by decoupling their language generation model from a reasoning agent, in which the language generation model is conditioned on the decisions or actions made by the underlying strategic agents.
Lastly, researchers have used pragmatic reasoning to improve text generation of image descriptions or instructions \citep{andreas-klein-2016-reasoning, fried-etal-2018-unified}.
Our work differs from prior work in that previously, language models were used to to ``reflect'' or ``manifest'' the actions being made by their counterpart reasoning agents.
On the other hand, our task is to provide strategically accurate \emph{analyses} about the actions that are taking place.

\section{Task description and data}
\label{sec:task}

In this section we provide a formal definition of our task, followed by details about our data.

\subsection{Task: Chess Commentary Generation}
\label{subsec:task_chess_comm_gen}
The goal of chess commentary generation is to provide a description or analysis of moves made in a chess game.
Analyses include suggesting alternative or future moves, or explaining the quality of a move.
The strategic nature of chess makes our task interesting, in that it requires robust language modeling as well as strong symbolic reasoning.

Borrowing \citet{jhamtani-etal-2018-learning}'s notations, given a game-state $G$, move $M$, and corresponding commentary text $C$, we want to model $P(C|G, M)$.
\citet{jhamtani-etal-2018-learning} introduces 6 categories of commentary types: move descriptions, move quality, move comparisons, rationales, contextual, and general.\footnote{See appendix for definitions and examples of each category.}
In this work we focus on the first three categories.\footnote{Our approach is generalizable for all categories. However, some categories include information that a chess engine does not provide, such as the historic context of a move. A separate expert system (chess database) could be substituted, but we leave this for future work.}

\subsection{Data}
\label{subsec:data}

Following \citet{jhamtani-etal-2018-learning}, we use commentary data from an online forum,\footnote{gameknot.com} where community members self-annotate chess games at a move-by-move granularity.
We collect a set of 373,919 triplets in the form of $(G, M, C)$.

\begin{figure*}[t]
  \centering
  \includegraphics[clip, trim=0.1cm 13.4cm 0.0cm 1.8cm, width=0.91\textwidth]{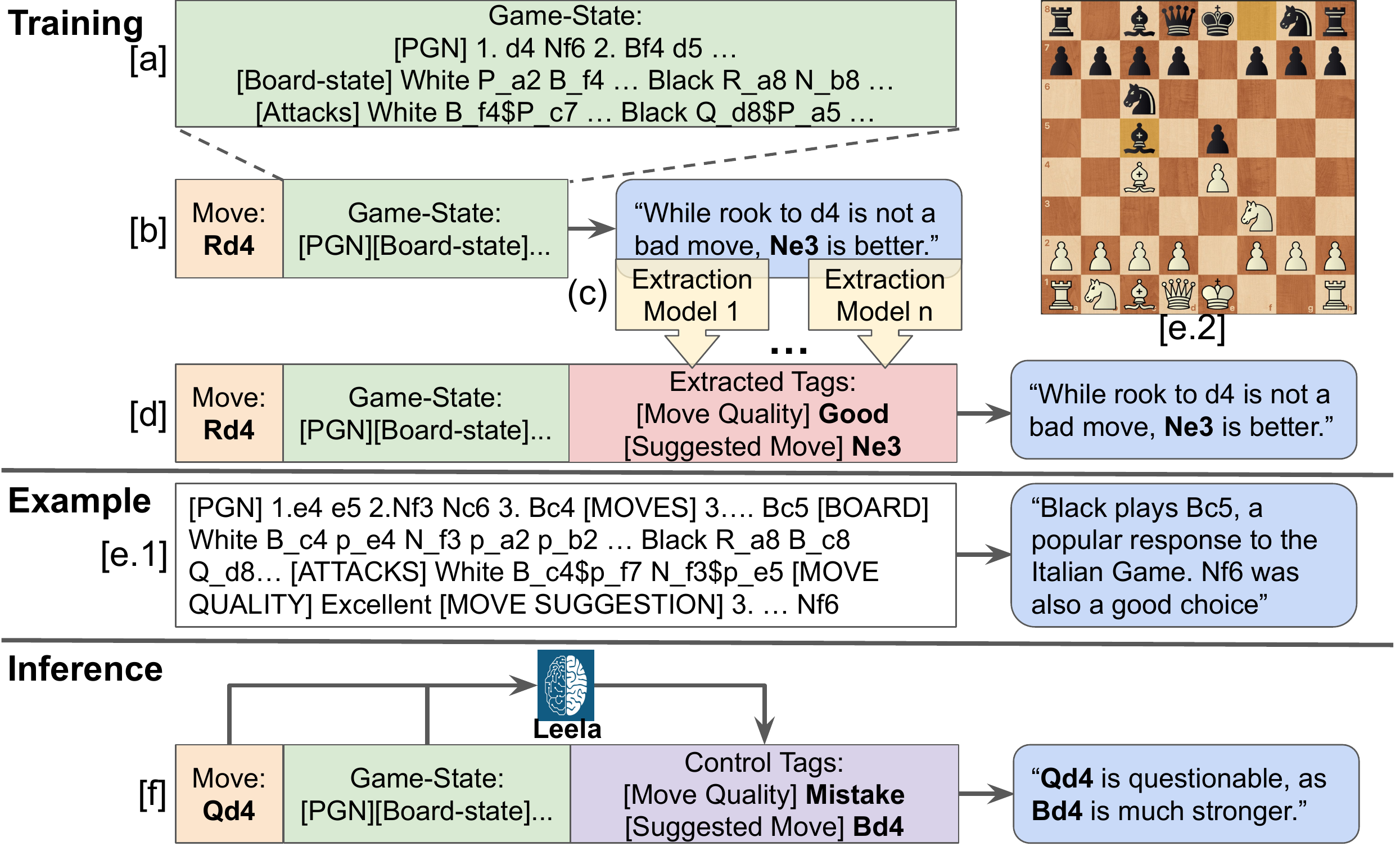}
  \caption{\label{fig:methodology}
\textbf{Training and inference pipeline of our commentary generation model.}
We build a representation of the game-state [a] and moves [b] and extract tags from the commentary using tag-extraction models [c].
The extracted tags provide training signals that align with the commentary text [d], thus giving us a controllable interface to use during inference.
Once a controllable commentary generation model is trained, we use Leela to generate tags to control our commentary generation [f]. See \autoref{sec:methods} for a more complete description of these methods.
}
\end{figure*}

\subsection{Additional pre-training data}
\label{subsec:pretrain_data}
In addition to commentary data, we use data from chess forums to pre-train ``tag-extraction'' models, or models used to extract tags from the training data on which to condition the commentary generation model (\autoref{subsubsec:methods_tag_extraction}).
We build question and response pairs using each of the following forums:

\paragraph{Chess.com} Chess.com has a dedicated forum for community members to post their games and ask for feedback.
Feedback is often provided both at a strategic high-level and for specific moves.
We use all question and answer pairs in which the question contains a chess game, resulting in 11,657 pairs.\footnote{We have received explicit permission from Chess.com to use this data.}

\paragraph{Chess StackExchange} StackExchange contains question and answer pairs pertaining to chess. Questions contain up to 5 categorical tags, while responses have arbitrarily many up or down votes.
We filter out irrelevant tags and non-positive up-vote responses, resulting with 11,549 pairs.

\paragraph{r/chess Subreddit} The subreddit r/chess contains multiple threads regarding chess.
We use a pre-existing Reddit dataset \citep{baumgartner2020pushshift} extracted by a third party and made available on PushShift\footnote{https://files.pushshift.io/reddit/}.
Of that data, we use threads in which the response post contains chess move notation or other game-specific text (see appendix).
We use a total of 32,826 dialogues.

Across all three datasets, any personally identifiable information (PII) was scrubbed.
\paragraph{} Given such question and response pairs, our pre-training task (\autoref{subsec:training}) is to generate the response given the question as input.
For our Chess.com data, we also include the game-state within each question as part of our input, where the game-state is represented using the PGN representation to be described in \autoref{subsubsec:methods_conditioned_comm_gen}.

\section{Methodology}
\label{sec:methods}

In this section we first provide an overview of our approach and subsequently explain each step.

\subsection{Overview}

Our approach is visualized in \autoref{fig:methodology}. Given training data in the form of $(G, M, C)$ (\autoref{fig:methodology} [b]), we first build a token representation of game-state $G$ ([a]).
We then extract various ``tags'' $T$ from commentary $C$ that describe the game-state using tag-extraction models ([c]). 
Tags are factual information regarding the game-state that is mentioned in the commentary, such as quality of a move or alternative lines of moves.
The goal is for the extracted tags $T$ to provide signals that are consistent with the semantics of the commentary text, such that they can be used as control codes by the commentary generation model.
The commentary generation model is then trained to be conditioned on the extracted tags: $P(C|G, M, T)$ ([d]).
An example of our task can be found in [e.1], which depicts [e.2].
Once our controllable commentary generation model is trained, during inference, a new game-state $G'$ is passed into a chess engine, which provides new tags $T'$ to control our commentary generation ([f]).
Details regrading each step are provided below.

\begin{table*}
\small
\centering
\renewcommand{\tabcolsep}{3pt}
\begin{tabular}{lllccc}
    \toprule
        Tag Type & Tags (or Examples) & Extraction Model & \# Annotations & Validation Accuracy$\dag$ & Test Accuracy$\dag$ \\
    \midrule
    
    Commentary Type & \makecell[l]{Move Description, Move Quality,\\ Comparative, etc.} & Classification & 1800 & 0.805 & 0.725  \\
    \midrule
    Move Quality & \makecell[l]{Excellent, Good, \\Inaccuracy, Mistake, Blunder} & Classification & 610 & 0.827 & 0.667 \\
    \midrule

    Suggested Moves &  ``Ne5'', ``Qxd4 Bg7'' & Generative & 830 & 0.806 & 0.674 \\
    \midrule

    \makecell[l]{Pronouns, \\Proper Nouns} & ``Her'', ``Carlsen's'' & \makecell[l]{Named Entity \\ Recognizer} & N/A & N/A & N/A \\
    \midrule

    Length & Short, Medium, Long & Heuristic & N/A & N/A & N/A \\
    \bottomrule
\end{tabular}
\caption{\label{tab:tag_extraction}
\textbf{Control tags extracted from commentary data.}
$\dag$ For classification models, we use F1 Scores. For generative models, we use token exact match scores.
}
\end{table*} 

\subsection{Training}
\label{subsec:training}

\subsubsection{Extracting tags}
\label{subsubsec:methods_tag_extraction}

First we identify tags $T$ to extract from our commentary training data.
For each tag type, we build a tag-extraction model $P(T|C)$, then annotate our commentary generation data $P(C|G, M)$ as $P(C|G, M, T)$, giving us finer control over the text generation task.
We extract 5 types of tags: ``commentary type'', ``move quality'', ``suggested moves'', ``pronouns/proper nouns'', and ``length''.

It is critical that the extracted tags are \emph{consistent} with the semantics of the commentary text. 
These tags provide the commentary generation model explicit signals to learn meaningful patterns between game-state $G$ and commentary $C$ during training.
For this reason, for the first 3 tags, we build tag-extraction models using BART \citep{lewis-etal-2020-bart} that is further pre-trained using the data described in \autoref{subsec:pretrain_data}. 
We then fine-tune each BART model using manually annotated samples as described below. 

\paragraph{Commentary Type} Commentary Type tags indicate which of the six categories of commentaries from \citet{jhamtani-etal-2018-learning} that a training example belongs to. 
To train a commentary type extraction model, we manually label 1,800 random samples of commentaries with one of the six categories.\footnote{We demonstrate our approach on 3 of the 6 categories (move description, move quality, move comparison), but our approach can be applied to the remaining categories using additional tags. Because we extract all 6 categories, our model technically can commentate regarding the remaining categories, but will likely be incorrect without additional tags.}
We then train a classifier by adding a linear classification layer to the last decoder layer of our pre-trained BART model.
The pre-trained weights for BART are frozen, and only the linear layer is learned.\footnote{Details regarding training is provided in the Appendix.}

\paragraph{Move Quality} Move Quality tags indicate how good or bad a move is.
Chess communities use a standard set, consisting of excellent, good, inaccurate, mistake, and blunder.
We manually annotate 610 commentaries, with an additional class ``None'' if the commentary does not refer to move quality.
We train a similar classifier as described above.

\paragraph{Suggested Moves} Suggested Move tags indicate alternative lines of moves that are mentioned in the commentary.
We manually label 830 commentaries, annotating the suggested moves $S$ using standard algebraic notation (SAN) to create ($C$, $S$) pairs.
$S$ is given the value ``None'' if there is no move being suggested.
Unlike the previous two classification models, here we use a generative model.
Namely, given ($C$, $S$), we fine-tune our pre-trained BART model to generate $S$ given $C$. 

\paragraph{Pronouns, Proper Nouns} In practice, we observe two additional tags that improve our commentaries.
Because our commentary data is from a community forum, the text includes multiple perspectives, including personal pronouns and proper nouns (grandmasters).
To generate a consistent perspective that contain neither, we tag our commentary data with pronouns and proper nouns that appear in each commentary.\footnote{We build an allow list of proper nouns that should be excluded, such as opening names (i.e., ``Ruy Lopez'')}
We use SpaCy's named entity recognizer\footnote{\url{https://spacy.io/api/entityrecognizer}} to extract these tags.

\paragraph{Length} We notice a correlation between the length of the generated commentary and its quality.
Short ones are often uninteresting (``White captures.''), while longer ones are more prone to including an error.
To control the length of our outputs, we tag our data with ``[short]'', ``[medium]'', or ``[long]'', depending on the commentary's length.\footnote{We use an arbitrary heuristic of 7 and 20 tokens as cut off points to assign length tags.}

\paragraph{}\autoref{tab:tag_extraction} summarizes our tag-extraction models.
Details for training are provided in the Appendix.

Using our tag-extraction models, we convert our task from $P(C|G, M)$ to $P(C|G, M, T)$.
Conditioning on tags $T$ provides an ``interface'' for symbolic reasoning systems to use.
Namely, during inference, given a new game-state $G'$ we use chess engines to derive new tags $T'$ to control our commentary generation model (\autoref{subsec:methods_inference}).

Our approach is akin to a line of work that similarly decouples the reasoning component from a language model \citep{doi:10.1126/science.ade9097, he-etal-2018-decoupling, pmlr-v80-yarats18a}.
Our task differs in that prior work uses language models to ``manifest'' the actions made by their underlying reasoning agents.
Our task requires generating strategically accurate analyses about actions that occur.

\subsubsection{Training a controllable commentary generation model}
\label{subsubsec:methods_conditioned_comm_gen}

For commentary generation, we train a BART model on the task of $P(C|G, M, T)$, in which $G$, $M$, and $T$ are represented using text.
Game-state $G$ is represented 3 ways: portable game notation, board-states, and attack-states (\autoref{fig:methodology} [a]).
An example of our input representation is shown in \autoref{fig:methodology} [e.1], which depicts the game-state in [e.2].

\paragraph{Portable Game Notation} 
PGN is a standard text format that tracks the history of moves played in a game.
Each move is notated using standard algebraic notation (SAN).
\citet{toshniwal2021chess} and \citet{demeter-downey-2021-whos} show that language models can track game-states for chess using move histories .  
Thus we include PGN representations, but also because commentaries often use SAN. 

\paragraph{Chess pieces}
Secondly, we enumerate the pieces on the board for each player.
For each player, we represent each piece by the piece-type, followed by their current square, separated by an underscore (``White R\_c5'').
The underscore is added to differentiate from the SAN tokens (``Rc5'').

\paragraph{Attacks}
Lastly, \citet{jhamtani-etal-2018-learning} demonstrate that featurizing on-going attacks on the board is helpful for commentary generation.
We represent all attacks by listing the attacking piece, followed by the attacked piece, separated by a special token (``\$'') (``White R\_a1\$P\_a2'').

\paragraph{} The game-state representations are concatenated with the move $M$ being played, and any conditional tags previously extracted (\autoref{fig:methodology} [d]).
Moves are represented using SAN.
Tags are preceded by special tokens to indicate the type of tag (``[Move Quality] Good'').
\autoref{fig:methodology} [e.1] shows an example input representation of the game-state in [e.2].

Given such input representation, we train BART to generate commentaries.
Details for training and hyperparameters are provided in the Appendix.

\subsection{Inference}
\label{subsec:methods_inference}

After training our commentary generation model, during inference, we use a chess engine to derive new tags to control our commentary generation.
We use Leela\footnote{https://lczero.org/}, an AlphaZero \citep{alphazero} variant agent that plays chess at a super-human level.
Concretely, given a new game-state $G'$ and move $M$, we use Leela to derive new tags $T'$, which in turn controls the generation of commentaries.
We explain how we derive each tag below.

\paragraph{Commentary Type} Commentary Type tags are simply selected by the user depending on what kind of analysis they wish to generate.

\paragraph{Move Quality} Given game-state $G$ and move $M$, we derive three values using Leela: $\mathbf{P\textsubscript{M}}$: the probability of winning after playing move $M$, $\mathbf{M*}$: the optimal move at game-state $G$, and $\mathbf{P\textsubscript{M*}}$: the probability of winning had $M*$ been played.
We use the difference between $P\textsubscript{M*}$ and $P\textsubscript{M}$ to classify the quality of move $M$ using a pre-defined set of ranges.\footnote{This approach is similar to how Chess.com classifies move qualities: https://support.chess.com/article/2965-how-are-moves-classified-what-is-a-blunder-or-brilliant-and-etc. We use the same set of range values for classification.}
Note that this value ranges from 0 to 1, where 0 indicates that move $M$ was the best move.

Given a move quality classification, we format it the same way we did for training. (\autoref{fig:methodology} [f]).

\paragraph{Suggested Moves} To commentate on move suggestions, we simply use Leela's suggested moves $M*$ (``[Suggested Move] Ne4'').

\paragraph{Pronouns, Proper Nouns} Recall that any commentary training data that contain pronouns or proper nouns are tagged.
In order to generate commentary that contain neither, during inference, we always \emph{omit} any pronoun or proper noun tags.

\paragraph{Length} For inference, we use ``medium'' as our length tag to control the length of our outputs.

\section{Results}
\label{sec:experiments}

In this section we show the accuracy of our tag-extraction models, followed by experiments to assess how well we are able to control the commentaries with our tags.
Finally, we use human judges to compare the quality of our commentaries against that of baseline models.

\subsection{Tag-extraction models}
\label{sec:experiments_tag_extraction}

To build our tag-extraction models, we use a 80:10:10 split of our manually annotated training data for classification models and a 85:10:5 split for our generative model.
\autoref{tab:tag_extraction} shows the resulting F1 and token exact match scores.
High accuracy is important to ensure that we are extracting the correct relationship between each game-state and resulting commentary, which, in turn, allows us to better control the commentary generation model using these tags.
For each tag type, our test accuracy ranges from 67\% to 73\%, and we find that such accuracy is sufficient in providing us control of our commentary generation model, as shown below.

\begin{figure}[t]
  \centering
  \includegraphics[clip, width=0.98\columnwidth]{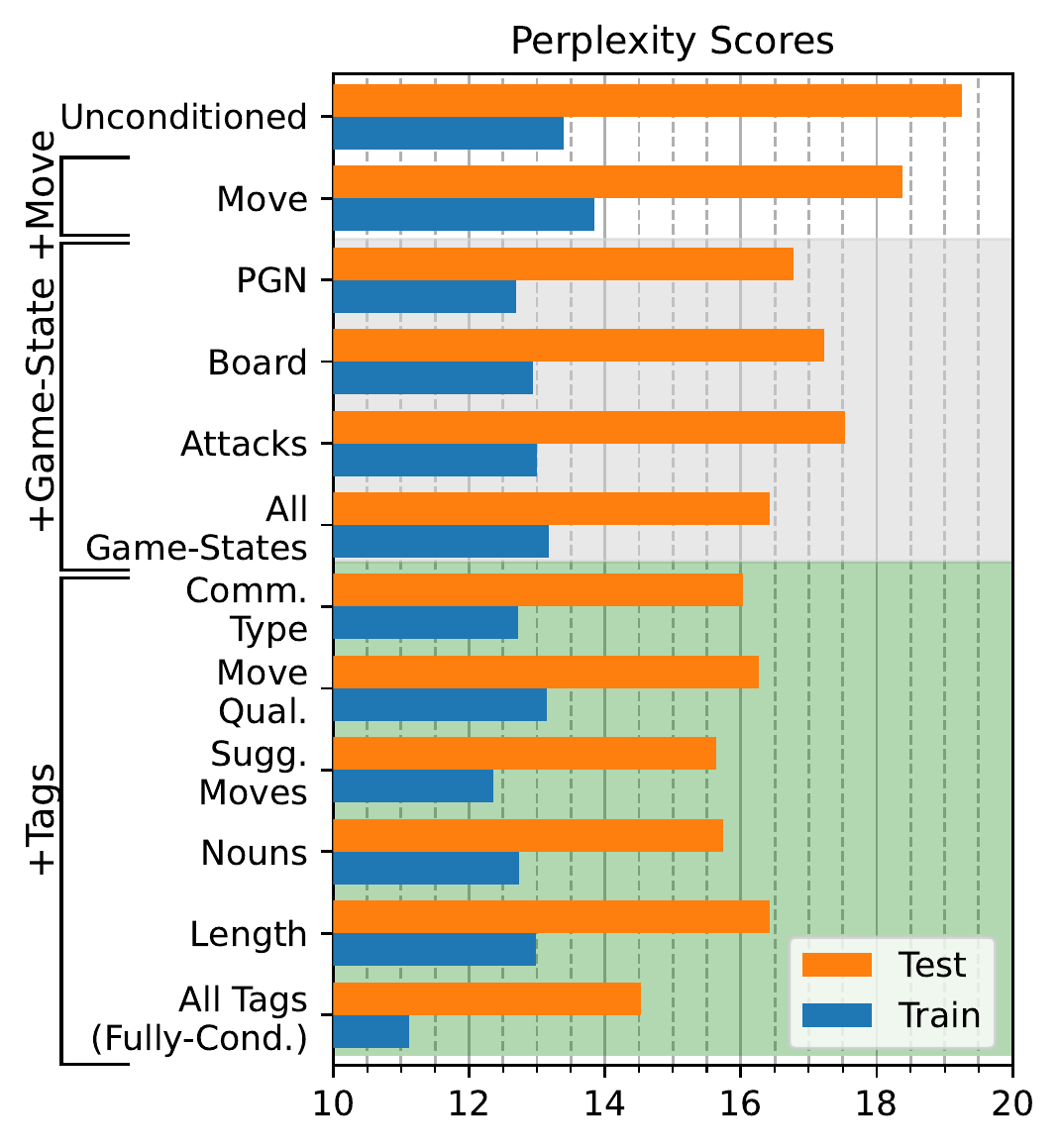}
  \caption{\label{fig:perplexity}
\fontdimen2\font=2pt
\textbf{Perplexity scores for different input representations.}
The y-axis indicate various input representations used by our commentary generation model.
Each grouping is cumulative of the previous grouping (i.e., +Game-State is in addition to Move, while +Tags is in addition to both Game-State (All) and Move).
Lower perplexity indicates that each tag contains useful signals for the model to learn patterns between the input representations and their corresponding commentary text.
}
\end{figure}

\begin{table}
\small
\centering
\begin{tabular}{llr}
    \toprule
       Commentary Type & Model & \makecell{Prefers Ours? \\ (By Majority Vote)} \\
    \midrule
    \multirow{3}{*}{Move Description}   & Unconditioned     & 55\% \\
                                        & Game-state        & 59\%  \\
                                        & Jhamtani et al.   & 72\%  \\
    \midrule
    \multirow{3}{*}{Move Quality}       & Unconditioned     & 65\% \\
                                        & Game-state        & 55\% \\
                                        & Jhamtani et al.   & 75\% \\
    \midrule
    \multirow{3}{*}{Comparative}        & Unconditioned     & 67\% \\
                                        & Game-state        & 54\% \\
                                        & Jhamtani et al.   & 72\% \\
    \bottomrule
\end{tabular}
\caption{\label{tab:human_eval}
\textbf{Human evaluation results.} We report the percentage of samples in which the majority vote from crowdworkers prefer our model compared to the three baselines. Our model outperforms all baselines, and is preferred over previous work (\citet{jhamtani-etal-2018-learning}) at least 72\% of the time for each commentary type.
}
\end{table} 

\begin{figure*}[t]
  \centering
  \includegraphics[clip, trim=0.25cm 5.7cm 0.2cm 0.1cm,width=0.98\textwidth]{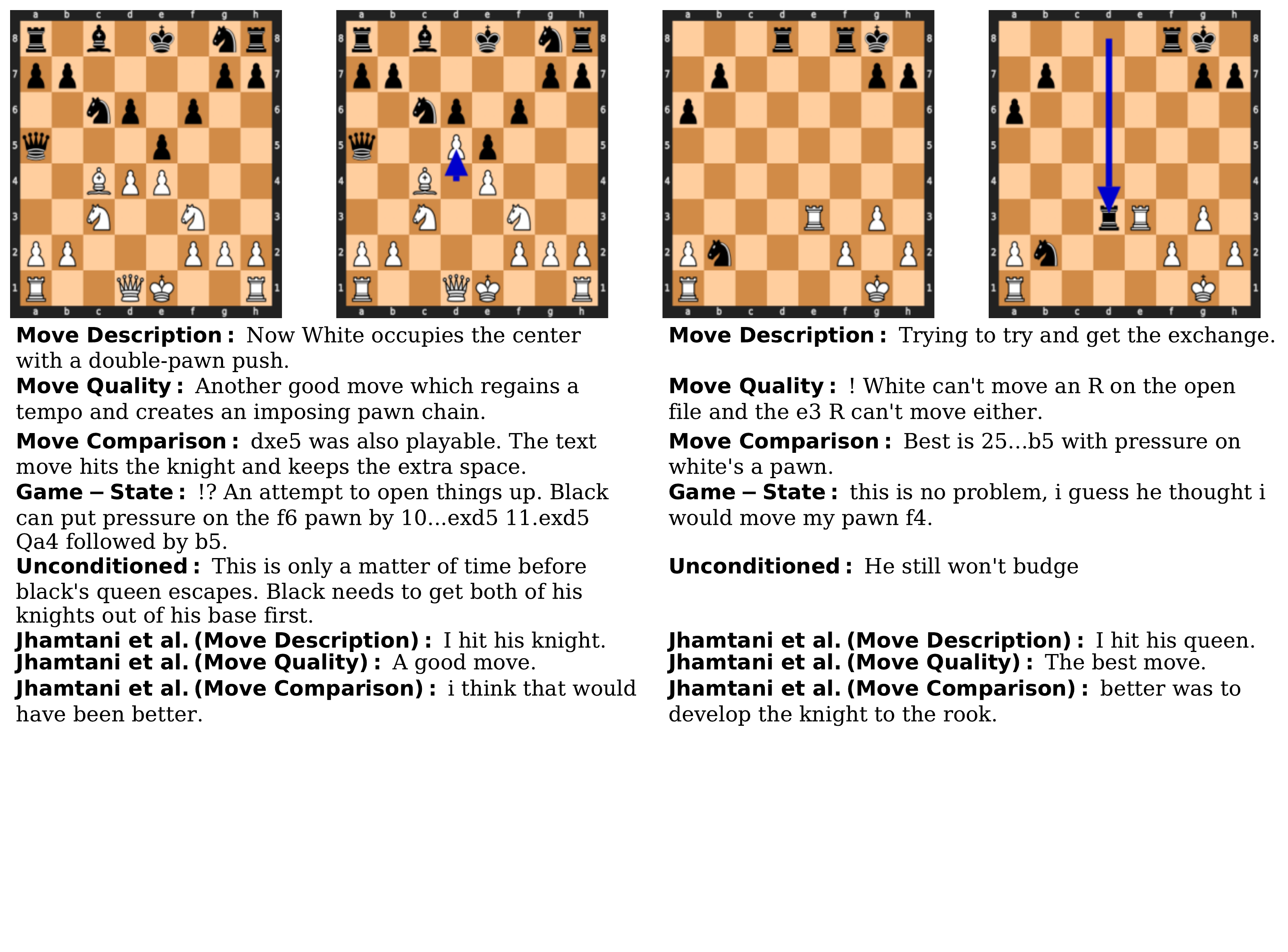}
  \caption{\label{fig:responses_main}
\textbf{Example commentaries generated by our models.}
The first three commentary types are from our fully conditioned model. 
More examples are provided in the Appendix.
}
\end{figure*}

\subsection{Perplexity}
\label{sec:experiments_ppl}

We expect a language model that learns a meaningful relationship between its input representation and corresponding commentary to see a reduction in perplexity.
\autoref{fig:perplexity} shows the drop in perplexity compared to different baselines, each of which are BART models trained on different input representations and tags, which we describe below.

\paragraph{Unconditioned} The simplest baseline is a completely unconditioned model.
Rather than grounding in the game state, we limit the input for this model to a single ``[Unconditioned]'' token.

\paragraph{Move} The next simplest model is one only conditioned on move $M$.
While it may learn to spell out a move (``Ne5'' \textrightarrow ``Knight to e5.''), it lacks grounding in the broader game state.

\paragraph{Game-State} ``Game-State'' refers to models that include a game-state representation in its input, in addition to move $M$.
\paragraph{Tags} ``Tags'' refers to models that include conditional tags in its input, in addition to using all 3 game-state representations and move $M$.
\paragraph{Fully-Conditioned} ``Fully-Conditioned'' refers to our model that uses all 3 game-state representations and all 5 conditional tags.

\paragraph{} As we provide our model more signals, in terms of both game-state representations and tags, we see a drop in perplexity, demonstrating its ability to learn meaningful patterns between our tags and corresponding commentaries.
Thus we expect such drop in perplexity to allow Leela to better control commentary generation during inference.

\subsection{Human evaluations}
\label{sec:experiments_human_eval}

Lastly, to evaluate our entire system, we conduct human evaluations to measure the quality of our generated commentaries compared to those of our baseline models.
Namely, we conduct A/B tests using Amazon Mechanical Turk.

Of our baseline models, we compare against \textbf{Unconditioned} and \textbf{Game-State (All)}.
We also compare against prior work: \citet{jhamtani-etal-2018-learning} is a LSTM-based language model trained on hand-crafted features, such as ongoing attacks.\footnote{While \citet{zang-etal-2019-automated} also suggest a chess commentary generation model, we are unable to reproduce their work and exclude their model from our experiments.}

\paragraph{Crowdworker task} We validate the chess knowledge of our crowdworkers using chess puzzles, in which crowdworkers are asked to find checkmating moves.\footnote{Details and screenshots are provided in the Appendix.}
Crowdworkers who do not solve them are not allowed to participate in our annotation tasks.

For our annotation task, we evaluate the three commentary types: ``Move Description'', ``Move Quality'', and ``Move Comparison''.
We present crowdworkers a \emph{sequence} of three game-states from a game.
Each game-state provides a pair of commentaries, one from our model and one from a baseline model.
We ask crowdworkers to select the system that overall generated better commentaries.
We also ask crowdworkers to optionally write the reasons for their choice (\autoref{sec:experiments_qualitative}).
An example of our tasks can be found in the Appendix.

Per commentary type and baseline comparison, we sample 100 random sequences.
Each sample is presented to three different crowdworkers, and we take the majority vote per sample.
\autoref{tab:human_eval} presents the percentage of samples in which crowdworkers prefer our model's commentaries.

The results show that our model strongly outperforms all baselines across all commentary types.
Most notably, for all commentary types, human judges prefer our model over \citet{jhamtani-etal-2018-learning} at least 72\% of the time.

\subsubsection{Qualitative analysis}
\label{sec:experiments_qualitative}

\autoref{fig:responses_main} demonstrates hand-selected examples of commentaries from each of our models, for each commentary type.\footnote{More examples are provided in the Appendix.}
We provide two takeaways. 
First, our fully-conditioned model generates much more accurate commentaries compared to baseline models.
For ``Move Quality'' and ``Move Comparison'', our model generates commentaries that are in agreement with chess engines.
However, not every aspect of its commentaries is correct.
For instance, in the second example of ``Move Quality'', while our model correctly commentates the move as good,\footnote{Chess communities often begin a commentary with ``!!'', ``!'', ``!?'', ``?'', or ``??'' to mark a move as excellent, good, inaccurate, mistake, or blunder.} its subsequent text is not correct (\autoref{sec:discuss}).

Interestingly, the reasons that crowdworkers provide for their selections align with our takeaways.\footnote{Random samples of reasons are provided in the Appendix.}
While crowdworkers often mention that our system is more accurate, a non-negligible number of workers also indicate that neither system is perfect.
Our system tends to err when commentating on multiple facets, which leaves more room for errors.

Secondly, crowdworkers often mention that our system understands good or bad plays, and that it provides correct move suggestions.
These were likely produced by the conditional tags from Leela.

\section{Discussion: limitations and challenges}
\label{sec:discuss}

We provide a brief discussion of limitations and challenges we observe.

\paragraph{Logical reasoning errors} 
Our model still occasionally generates commentaries with logical reasoning errors, which often correspond to information lacking in the input representation.
Although we leverage a symbolic reasoning engine to relieve our language model the burden of reasoning tasks, the current interface does not cover every possible logical deduction, such as future game-states. 
Building a more comprehensive interface with richer tag annotations is one possible solution.
Alternatively, learning reasoning skills for chess \citep{toshniwal2021chess, demeter-downey-2021-whos} without sacrificing the language model's strong natural language prior could be an interesting direction.

\paragraph{Human judgement of super-human systems}
Interestingly, crowdworkers sometimes disagree with our model's assessment, even when the commentaries are accurate, suggesting that crowdworkers sometimes disagree with Leela.
It is commonly known that the super-human behavior of chess engines are different from that of humans and thus difficult to interpret \citep{mcilroyyoung2020maia}.
While human judgement is often used as gold standard, it may be insufficient in evaluating future super-human capable systems.

       

\begin{figure}[h]
  \centering
  \includegraphics[clip, trim=0.3cm 0.2cm 0.0cm 0.2cm, width=0.96\columnwidth]{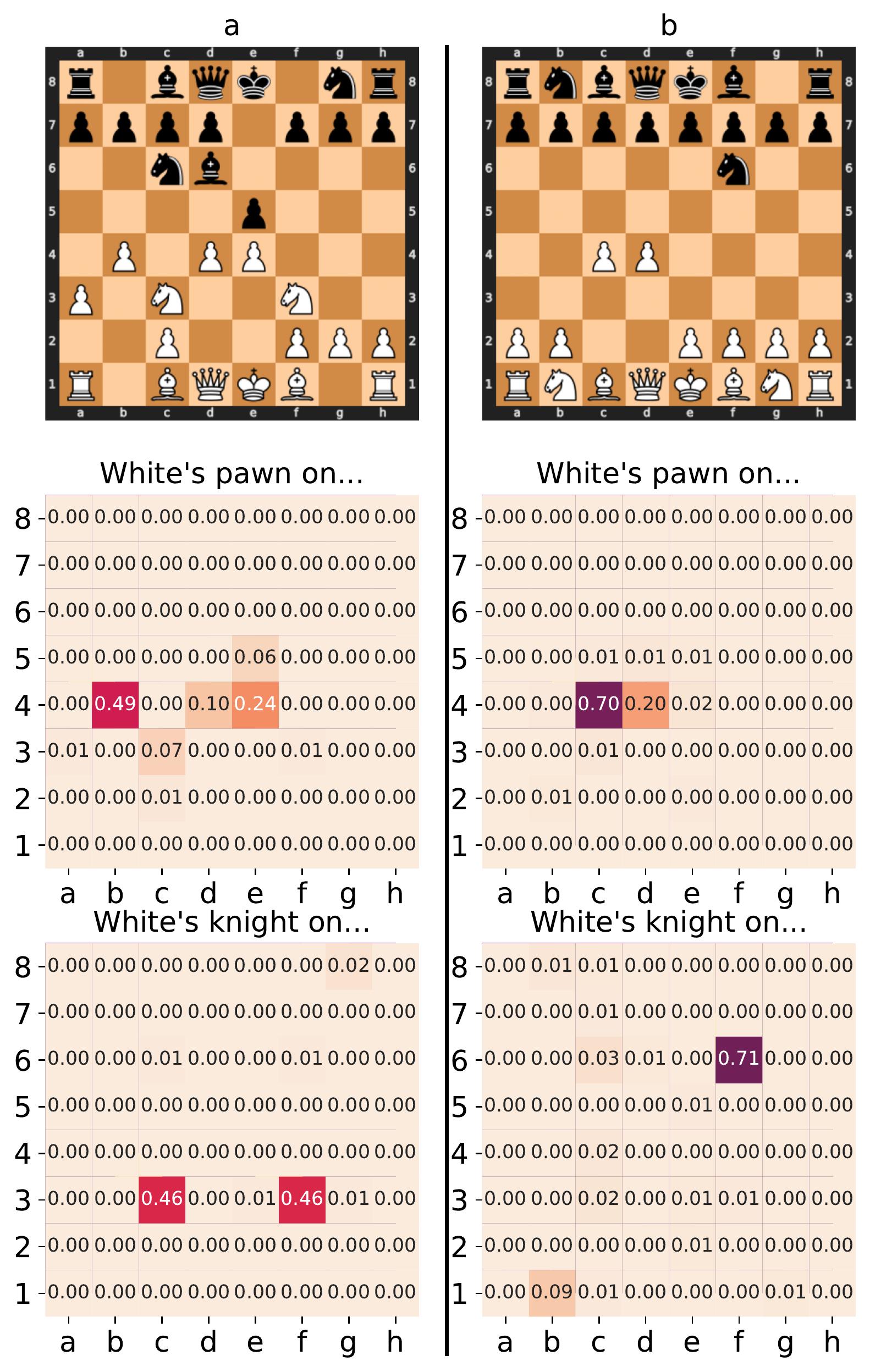}
  \caption{\label{fig:heatmaps}
\textbf{Prompted belief-states of our language model.}
This visualization can provide insights into our model's understanding of the board state.
}
\end{figure}

\paragraph{Grounding error analysis} 

Occasionally, our models generate commentaries that contradict the game state on which they are conditioned. 

To better understand this phenomenon, we use templated prompts to derive the model's belief-state about pieces on the board, i.e., we prompt the model with text like ``White's bishop on '', and compute the conditional likelihood for each of the 64 chess squares (e.g., ``e5''). See \autoref{fig:heatmaps} for an example visualization. Our results show that while our model occasionally makes mistakes, it exhibits the best understanding about pieces which were recently moved. Intuitively, this makes sense, as commentaries are most likely to discuss recently moved pieces. Experimental details, complete results, and further analysis are provided in \autoref{appx:heatmaps}.

\section{Conclusion}
\label{sec:conclusion}

In this paper, we demonstrate how to combine a symbolic reasoning system with a controllable language model to generate chess commentaries.
By extracting control tags from commentary data that share a relationship with the corresponding game-state, we provide our commentary generation model with explicit signals to learn meaningful patterns.
Conditioning this model on such signals provide an interface to incorporate a super-human chess engine during inference.
Namely, we use the outputs from Leela to control our commentary generation.
Our experiments show the utility of our tags, by demonstrating a drop in perplexity and that commentaries conditioned on Leela's tags are preferred by human judges.

\bibliography{anthology,custom}
\bibliographystyle{acl_natbib}

\clearpage

\appendix

\begin{table*}
\small
\centering
\renewcommand{\tabcolsep}{3pt}
\begin{tabular}{l|l|l}
    \toprule
    \textbf{Commentary Type} & \textbf{Description} & \textbf{Example} \\
    \midrule
    Move Description & Explicit or implicit description of the move & "Knight to e5." \\
    \midrule
    Move Quality & Quality of the move & "An inaccuracy for white as it leaves the rook hanging." \\
    \midrule
    Move Comparison & Comparison of multiple moves & "Developing the knight would have been preferred." \\
    \midrule
    Planning / Rationale & \makecell[l]{The rationale of the move in terms of \\ future gameplay, board advantages, etc.} & "Trying to force a queen's trade and prevent Bb5." \\
    \midrule
    Contextual & \makecell[l]{Description of the overall game state, such as \\ possibility of win vs. loss} & "White has the clear advantage." \\
    \midrule
    General & \makecell[l]{General information, such as information \\ about the players, tournament, or other remarks \\ irrelevant to the current game-state.} & "Game between Carlson and Hikaru." \\
    \bottomrule
\end{tabular}
\caption{\label{appx_tab:commentary_types}
Commentary Types
}
\end{table*} 

\section{Commentary Types}
\label{appx_sec:commentary_types}

We use the commentary types as introduced in \citet{jhamtani-etal-2018-learning}.
Their definitions and examples are provided in Table~\ref{appx_tab:commentary_types}.

\section{Cleaning Reddit Data}
\label{appx_sec:reddit}

Reddit data tends to be noisy.
In order to find question-answer pairs that are likely referring to a specific chess game, we exclude any threads that contain external links.
We then match the comment post against a set of patterns:

\begin{enumerate}
  \item PGN or SAN move notations.
  \item Referral to specific move numbers (ie., "move 10").
  \item chess "event" tokens.\footnote{exchange, castle, capture, blunder, mate, check, checkmate, discovered attack/en passant, fianchetto, gambit, pin, sacrifice, stalemate, threat, trap, variation}
  \item chess piece tokens.
\end{enumerate}

If any of the patterns are matched, we keep the comment post as the response, and use the previous thread (dialogue history) as context.

\begin{table}
\small
\centering
\renewcommand{\tabcolsep}{3pt}
\begin{tabular}{lc}
    \toprule
        Source & \# Data \\
    \midrule
        Chess.com & 11,657 \\ 
        Chess StackExchange & 11,549 \\
        Reddit r/chess & 32,826 \\
    \bottomrule
\end{tabular}
\caption{\label{tab:pretraining_data}
Pre-training Data.
}
\end{table} 

\section{Training Details, Hyperparameters}
\label{appx:hyperparams}

In this section we provide details regarding training our models, including hyperparameters that were searched.
We use ParlAI \citep{miller2017parlai} to train our models.
All of our models are based on a 4M parameter BART model.
We use Adam optimizers for each model.
Training was done on 8 NVidia V100 GPUs.

For commentary generation, we use a total of 373,919 samples of game-state and commentary pairs, which were split into 85:10:5 splits for training, validation, and testing.
For additionally pre-training our BART model, which is used by our tag-extraction models, we use a total of 56,032 samples of question and answer pairs, as summarized by Table~\ref{tab:pretraining_data}, which was also split into 85:10:5 splits.

For our commentary type extraction model, we use 1800 samples with 80:10:10 splits.
For our move quality extraction model, we use 610 samples with 80:10:10 splits.
For our suggested move extraction model, we use 830 samples with 80:15:5 splits.

For each of our generative models (pre-training, suggested move extraction, commentary generation), we minimize the validation perplexity during training.
For our classification models (commentary type extraction, move quality extraction), we minimize the validation loss.
Table~\ref{tab:hyperparams_pretrain} through ~\ref{tab:hyperparams_comm_gen} include the hyperparameters used to train our models.
The hyperparameter with the best performance is indicated as bold.

\section{Human Evaluations}
\label{appx:human_eval}

\begin{figure}[t]
  \centering
  \includegraphics[clip, trim=0cm 0cm 4.2cm 0cm, width=0.98\columnwidth]{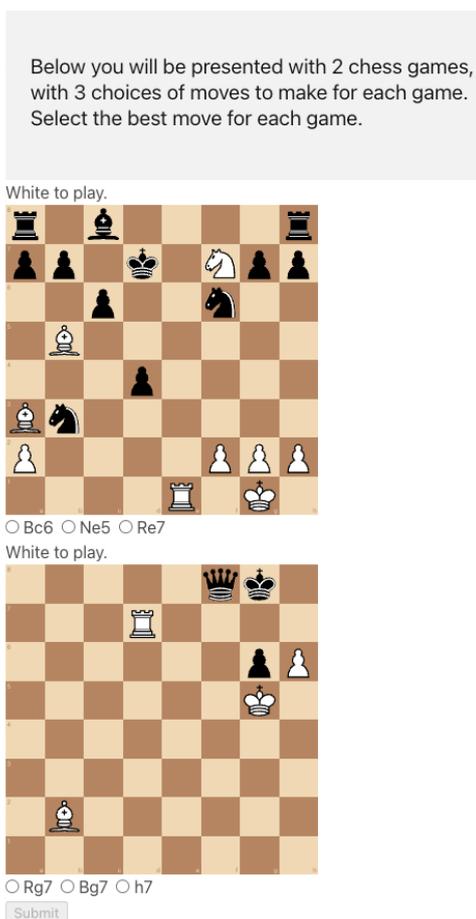}
  \caption{\label{fig:crowdsource_screen}
\textbf{Screenshot of our assessment task.}
Each puzzle presents 3 options of moves to play, in which one of them is a checkmating move.
Each move is presented in standard algebraic notation (SAN), to ensure that crowdworkers know how to parse SAN but also because SAN is commonly used in chess commentary.
}
\end{figure}

\begin{figure*}[t]
  \centering
  \includegraphics[clip, width=0.98\textwidth]{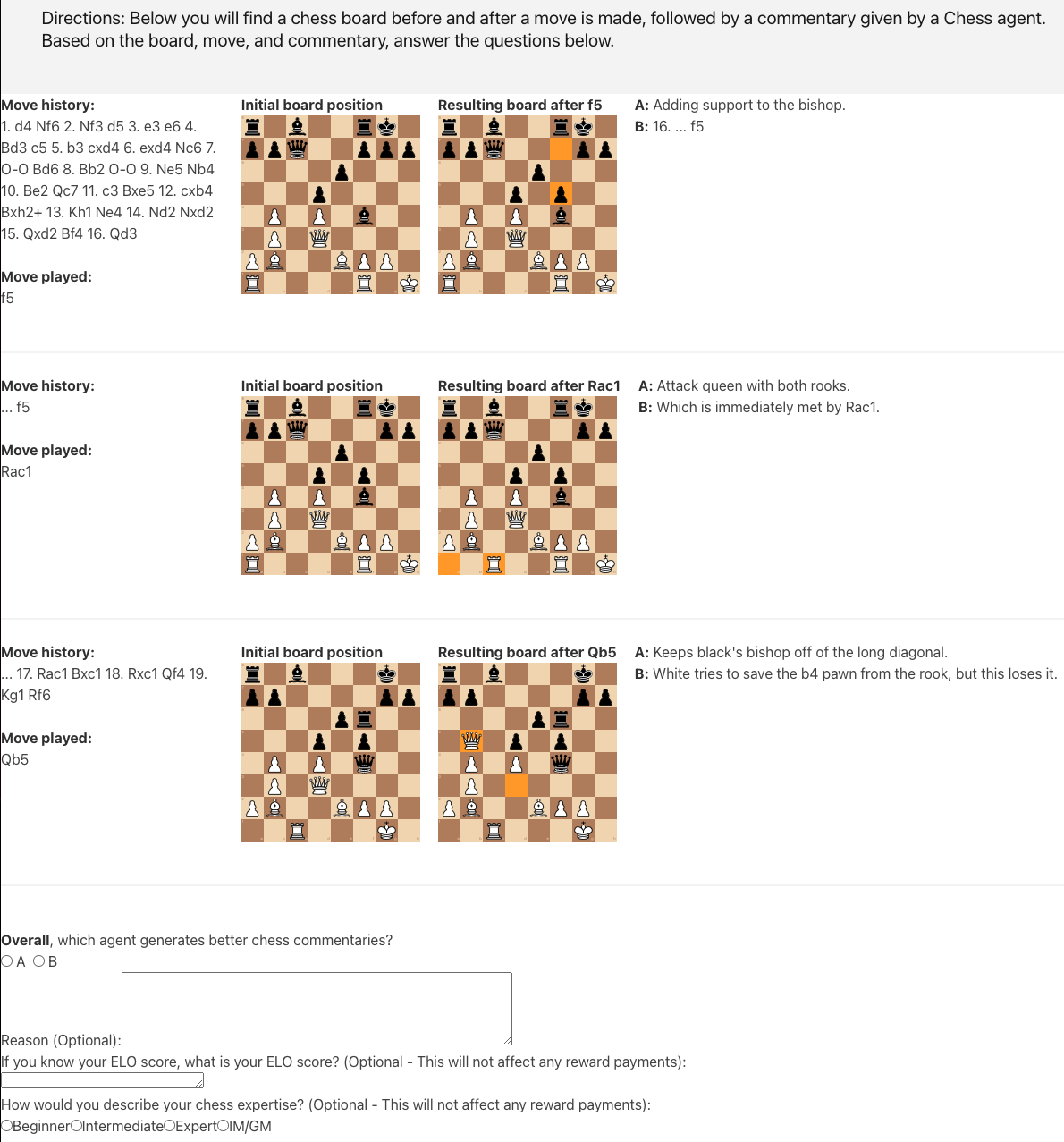}
  \caption{\label{fig:crowdsource}
Screenshot of our annotation task.
}
\end{figure*}

Figure~\ref{fig:crowdsource_screen} and Figure~\ref{fig:crowdsource} contain an example of our crowdsourcing assessment task as well as our main task.
Our assessment task consists of two random games, each followed by 3 choices of moves, one of which is a checkmate.
We ask crowdworkers to select the best move.
Each choice is given using standard algebraic notation (SAN), to ensure that crowdworkers know how to parse SAN, as SAN is typically used in chess commentary.\footnote{In SAN, checkmating moves include a \# symbol, which we remove.}


\subsection{Qualitative Analysis - Provided Reasons}
\label{appx:human_eval_reasons}

As part of our human evaluation study, we optionally ask each crowdworker to provide a reason for their selection.
In this section we provide random samples of reasons provided by crowdworkers.
Namely, for both cases in which a crowdworker prefers our model, as well as when they prefer a baseline model, we randomly sample 100 reasons and categorize them.
Table~\ref{tab:reasons_ours_i} and Table~\ref{tab:reasons_ours_ii} present reasons for human judges preferring our model, while Table~\ref{tab:reasons_baselines} presents reasons for preferring a baseline model.
Each table also includes the percentage of reasons that fall under each category, which include ``accuracy'', ``strategy'', and ``diversity, details, other''.
All other reasons that are not informative (e.g., ``I just like system A better.'') are categorized as ``Non-informative''.
Note that when human judges prefer our model, the number of times it is because of accuracy (36\% vs. 15\%) or strategy (14\% vs. 7\%) is much higher.

\section{Response Samples}
\label{appx:response_samples}

\begin{figure*}[t]
  \includegraphics[clip, trim=0cm 0cm 0cm 0cm, width=0.95\textwidth]{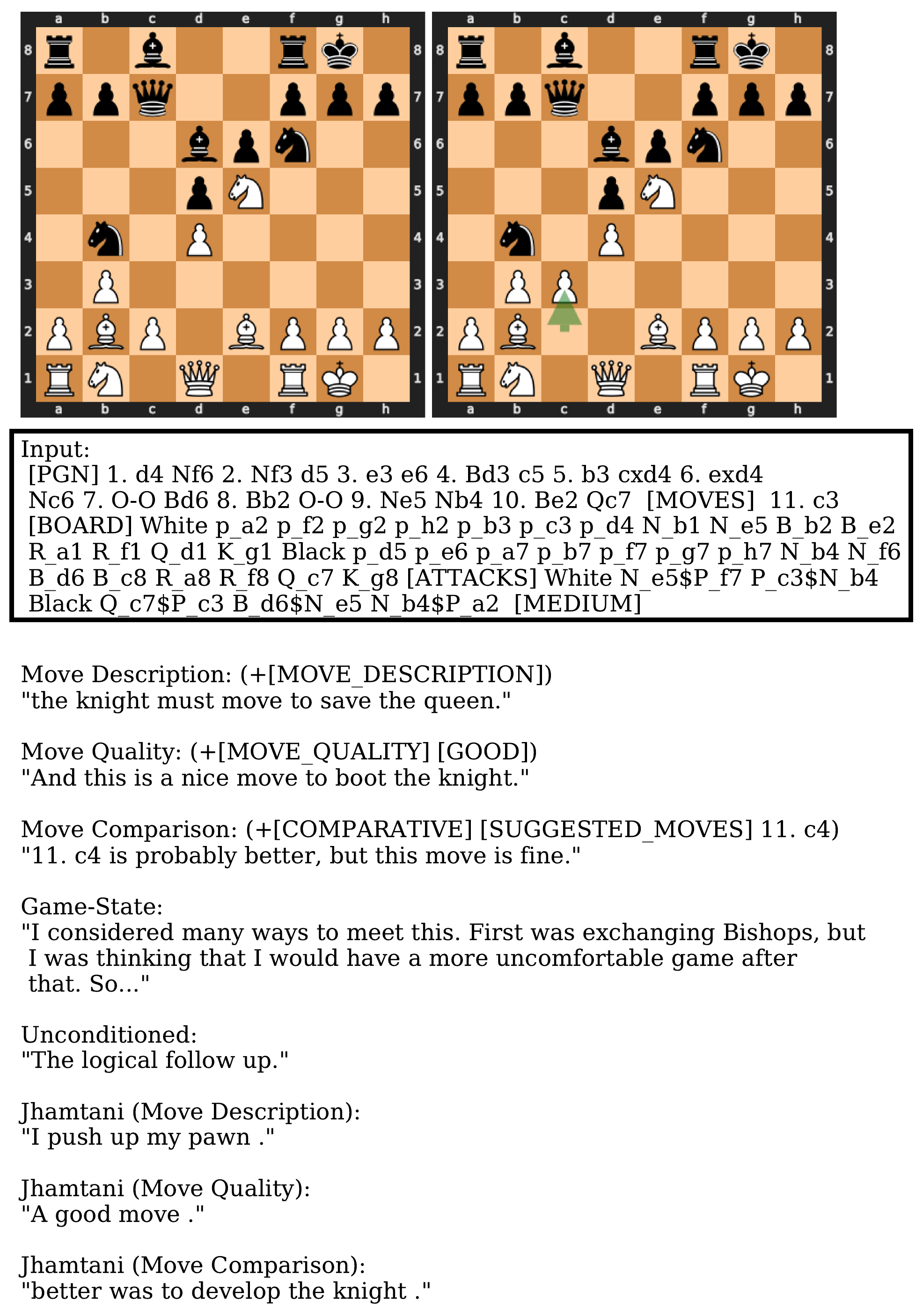}
  \caption{\label{appx_fig:response_samples_0}
Example of responses from our models.
The first three responses are from our fully conditioned model, in which we indicate what tags from Leela were used.
The generated responses from our fully-conditioned model seem reasonable for this example.
}
\end{figure*}

\begin{figure*}[t]
  \includegraphics[clip, trim=0cm 0cm 0cm 0cm, width=0.95\textwidth]{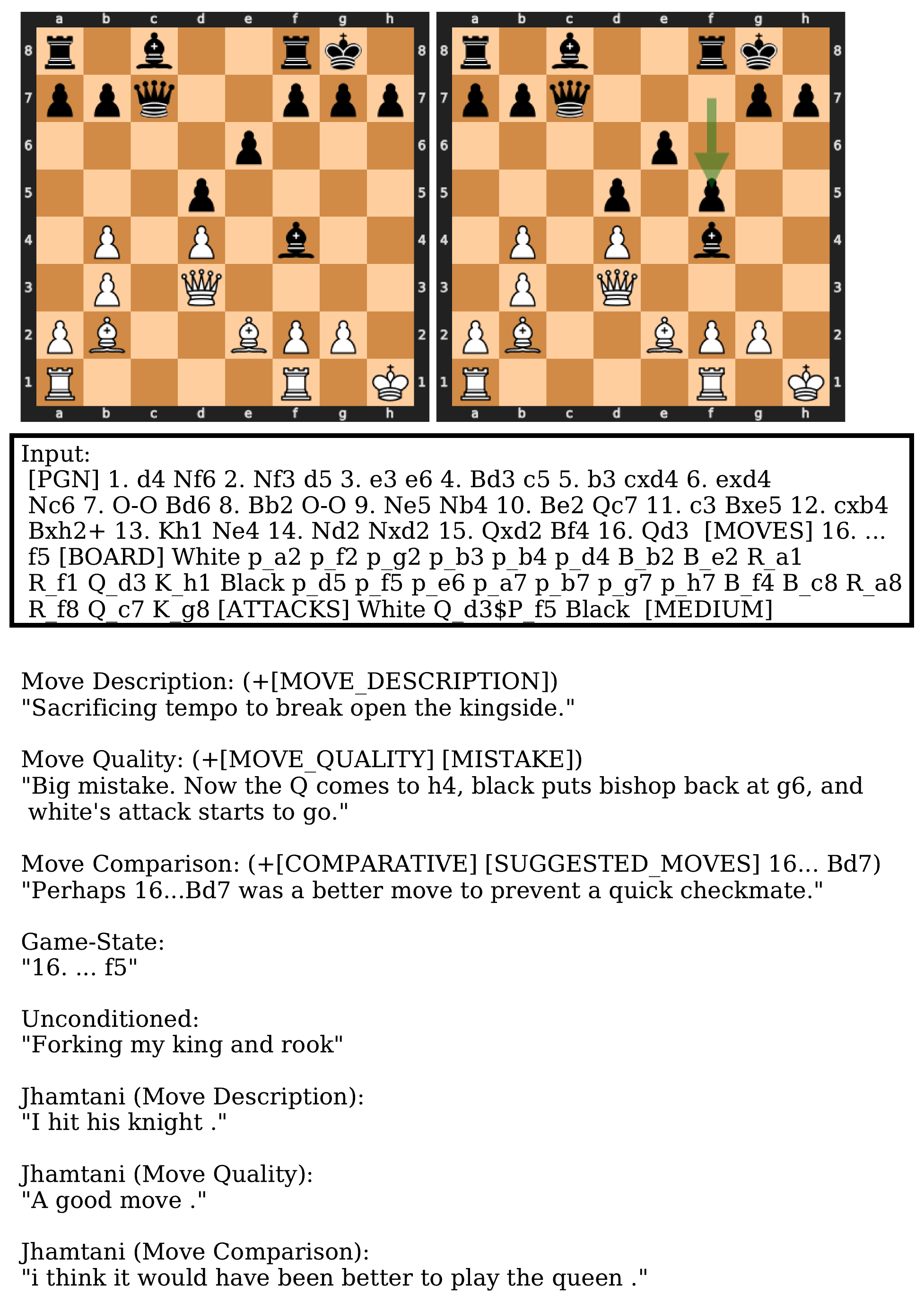}
  \caption{\label{appx_fig:response_samples_1}
For Move Quality, the model makes a logical error.
While the model correctly identifies the move as a bad mistake, it incorrectly identifies moves for the queen and bishop.
For Move Comparison, we also see an error.
While correctly commentating on the better move, the model mentions a quick checkmate that does not exist.
}
\end{figure*}

\begin{figure*}[t]
  \includegraphics[clip, trim=0cm 0cm 0cm 0cm, width=0.95\textwidth]{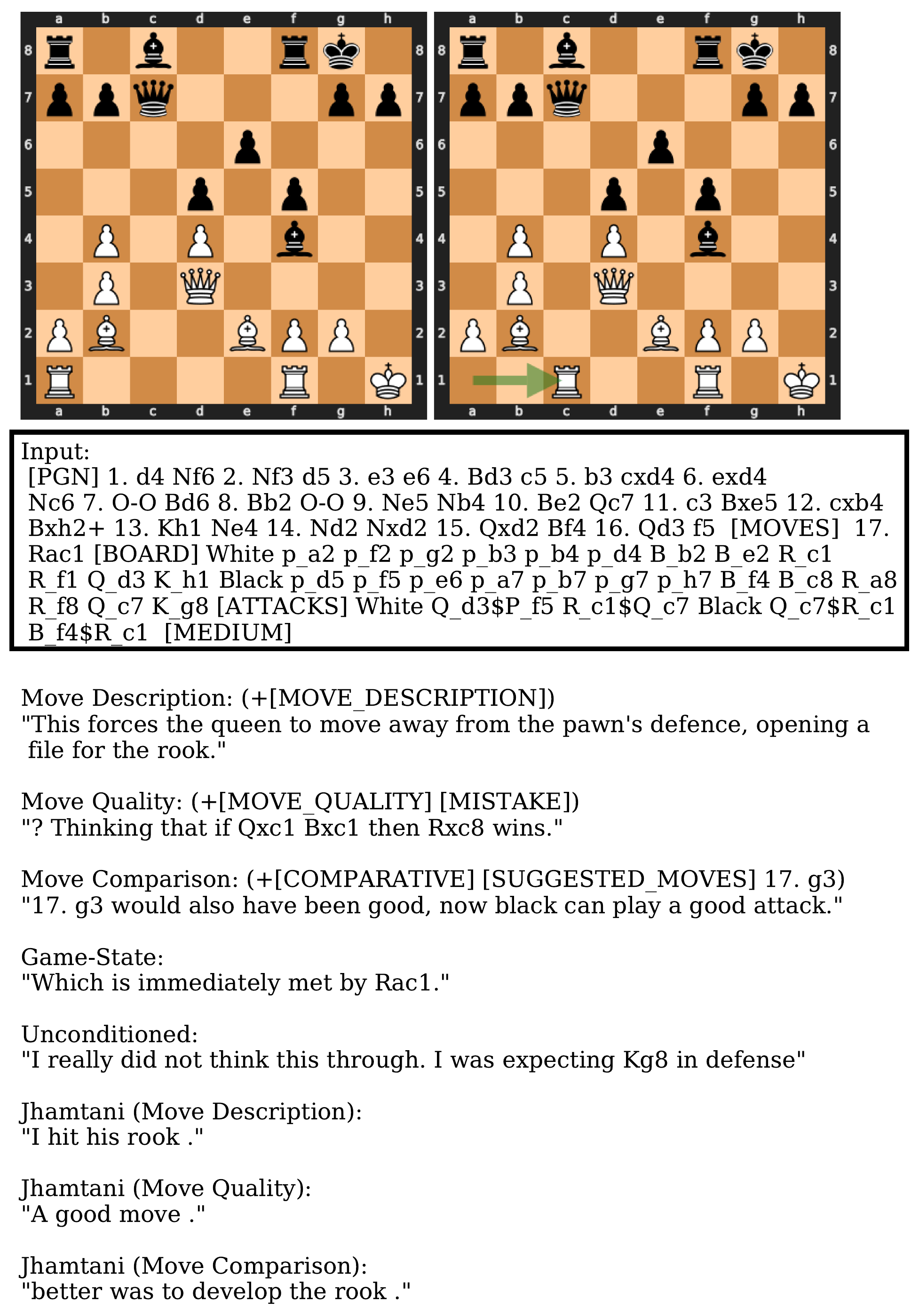}
  \caption{\label{appx_fig:response_2}
For Move Quality, the model correctly commentates that the move was a mistake by starting the commentary with "?" ("!!", "!", "!?", "?", or "??" are commonly used in the chess community to refer to a move as excellent, good, inaccurate, mistake, or blunder).
However, the following line of moves is incorrect.
All other responses from our fully conditioned model seem accurate.
}
\end{figure*}

\begin{figure*}[t]
  \includegraphics[clip, trim=0cm 0cm 0cm 0cm, width=0.95\textwidth]{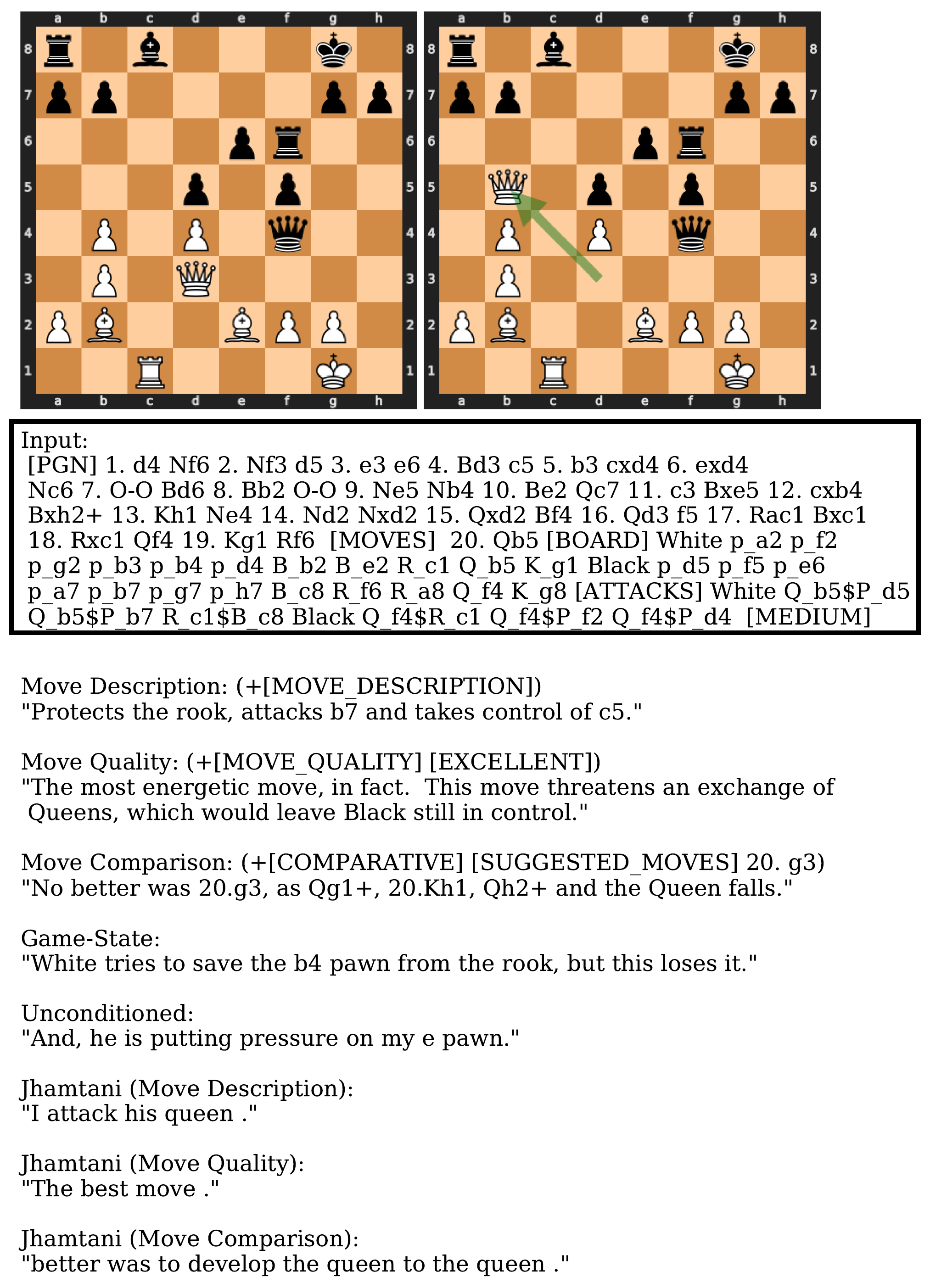}
  \caption{\label{appx_fig:response_samples_3}
For Move Description, the model commentates on 3 items -- protecting the rook, attacking b7, and controlling c5, of which 2 are correct.
Note that the incorrect information (protecting the rook) is not present in the input representation, while the other are.
For Move Quality, the model incorrectly commentates about exchanging queens.
For Move Comparison, the model incorrectly interprets the suggested move (20. g3) as an inferior move.
This occurs because our tag-extraction model for move suggestion does not distinguish alternative moves that are mentioned in the commentary text as a better move or inferior move to what was played.
Improving the tag-extraction model to distinguish the two cases would fix this issue.
}
\end{figure*}

\begin{figure*}[t]
  \includegraphics[clip, trim=0cm 0cm 0cm 0.2cm, width=0.95\textwidth]{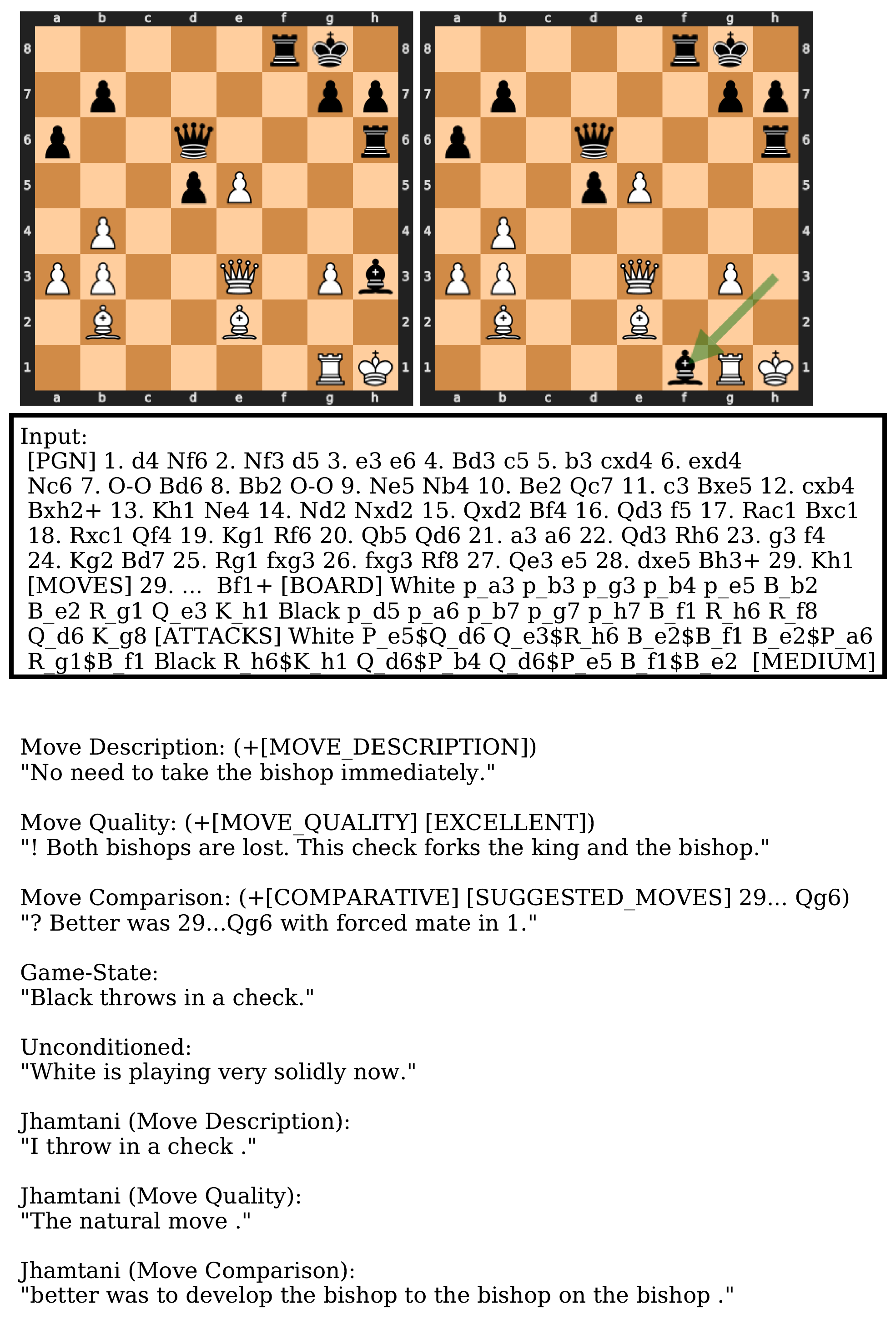}
  \caption{\label{appx_fig:response_samples_4}
For Move Quality, the model correctly commentates that the move was a good move.
Interestingly, after this move black now attacks the king and the bishop.
However, this is not a fork.
This may be a spurious pattern that the model has learned, in which when two simultaneous attacks appear after a move, the model has incorrectly learned it as a fork.
For Move Comparison, the model incorrectly commentates "mate in 1".
While this is incorrect, an additional "Mate in N" tag would allow us to control this behavior.
}
\end{figure*}

\begin{figure*}[t]
  \includegraphics[clip, trim=0cm 0cm 0cm 0cm, width=0.95\textwidth]{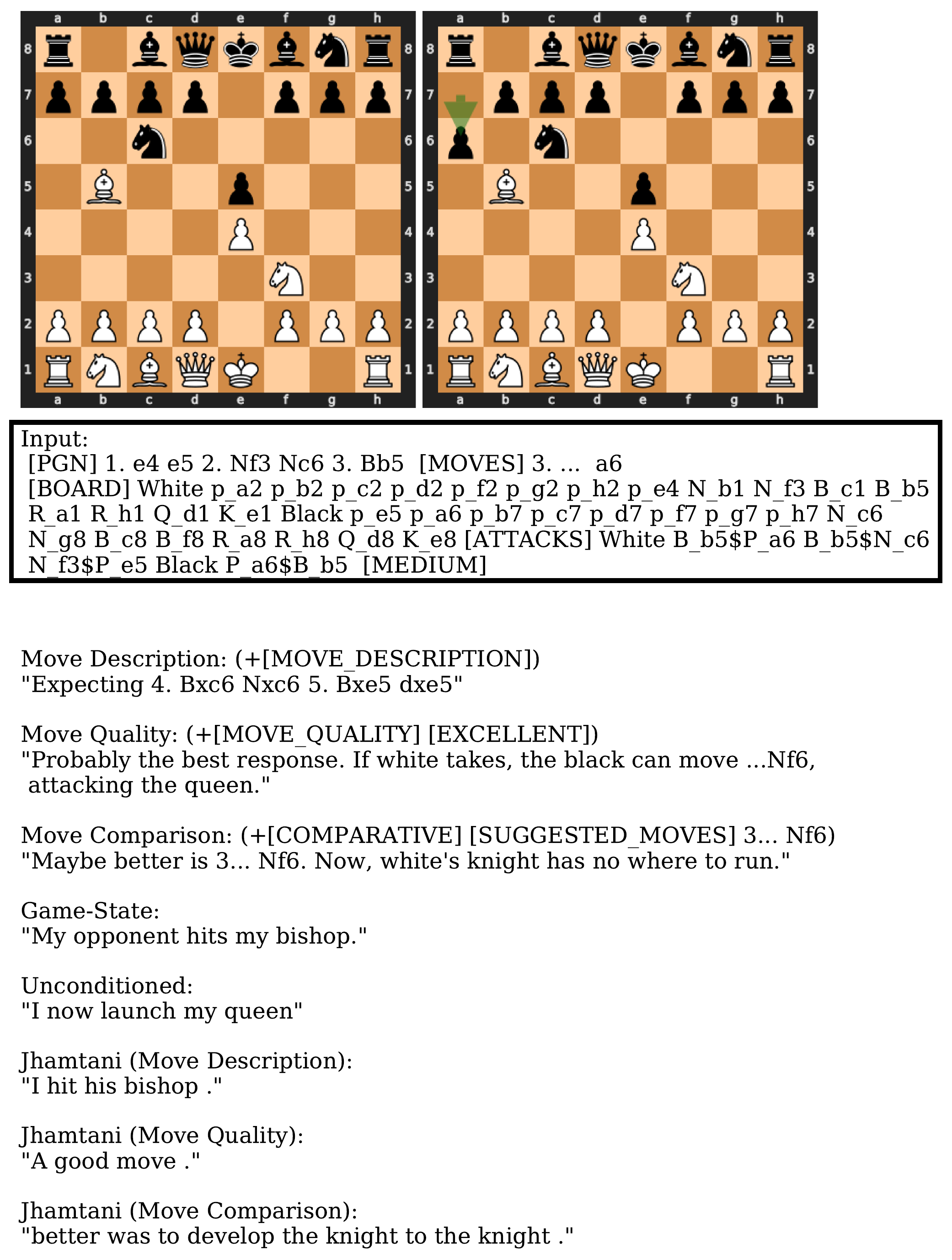}
  \caption{\label{appx_fig:response_samples_5}
For Move Quality, the model correctly commentates the move as the best move.
However, attacking the queen is incorrect.
Interestingly, although Nf6 is not mentioned in the input representation, the model commentates about such move, likely because in this current Ruy Lopez opening, Nf6 is a commonly seen move.
For Move Comparison, while the model correctly commentates about the best move, it contains a logical error by misidentifying white knight's status.
This would require the model to understand both valid moves per piece as well as future game-states, which are information not currently in the input representation.
}
\end{figure*}

\begin{figure*}[t]
  \includegraphics[clip, trim=0cm 0cm 0cm 0cm, width=0.95\textwidth]{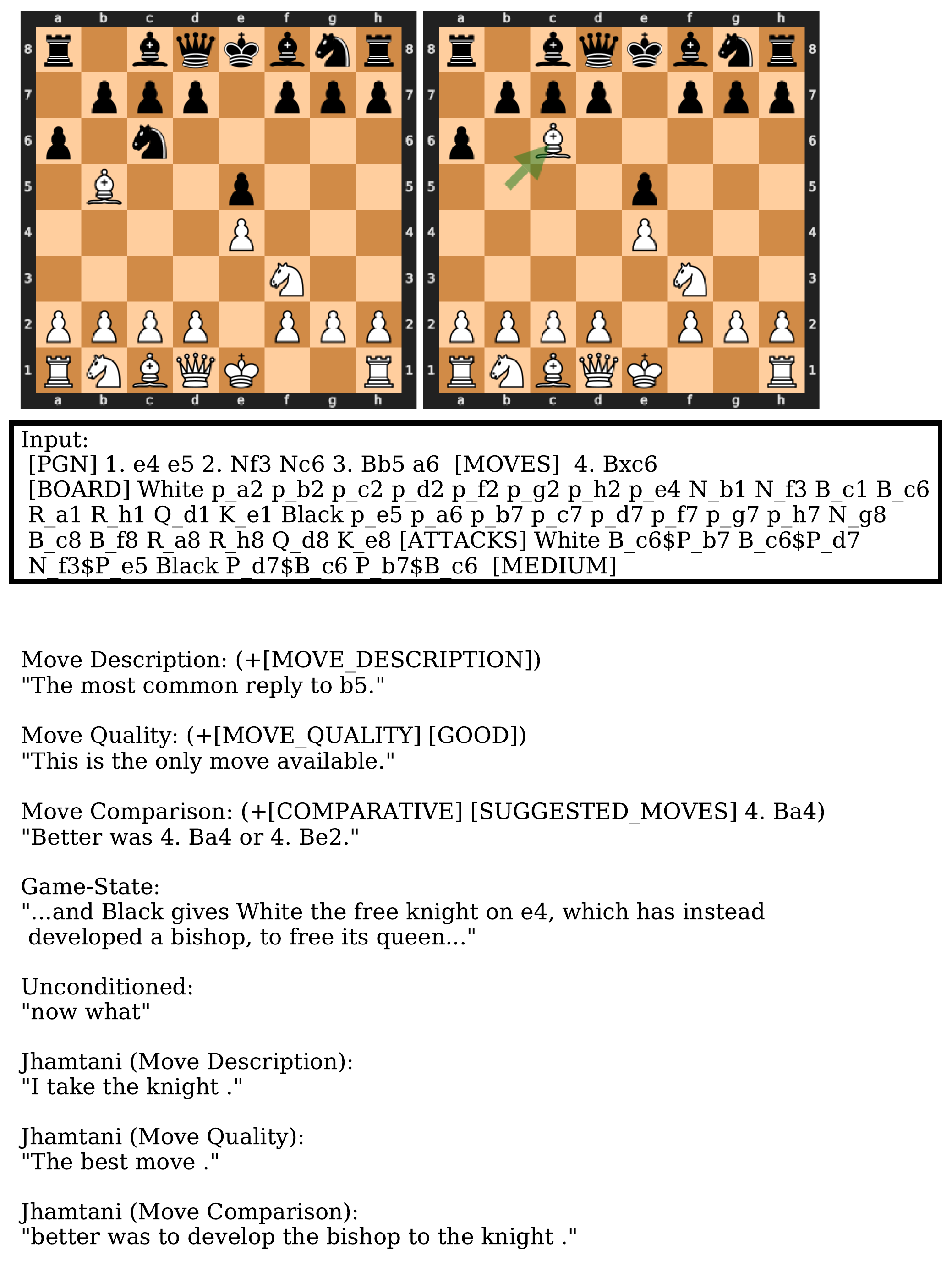}
  \caption{\label{appx_fig:response_samples_6}
All 3 responses from our model seem reasonable.
Interestingly, for Move Comparison, the model also identifies 4. Be2 as a good move, although such information is not present in the input representation.
This is likely because 4. Be2 is a commonly seen move in this opening.
}
\end{figure*}

\begin{figure*}[t]
  \includegraphics[clip, trim=0cm 0cm 0cm 0cm, width=0.95\textwidth]{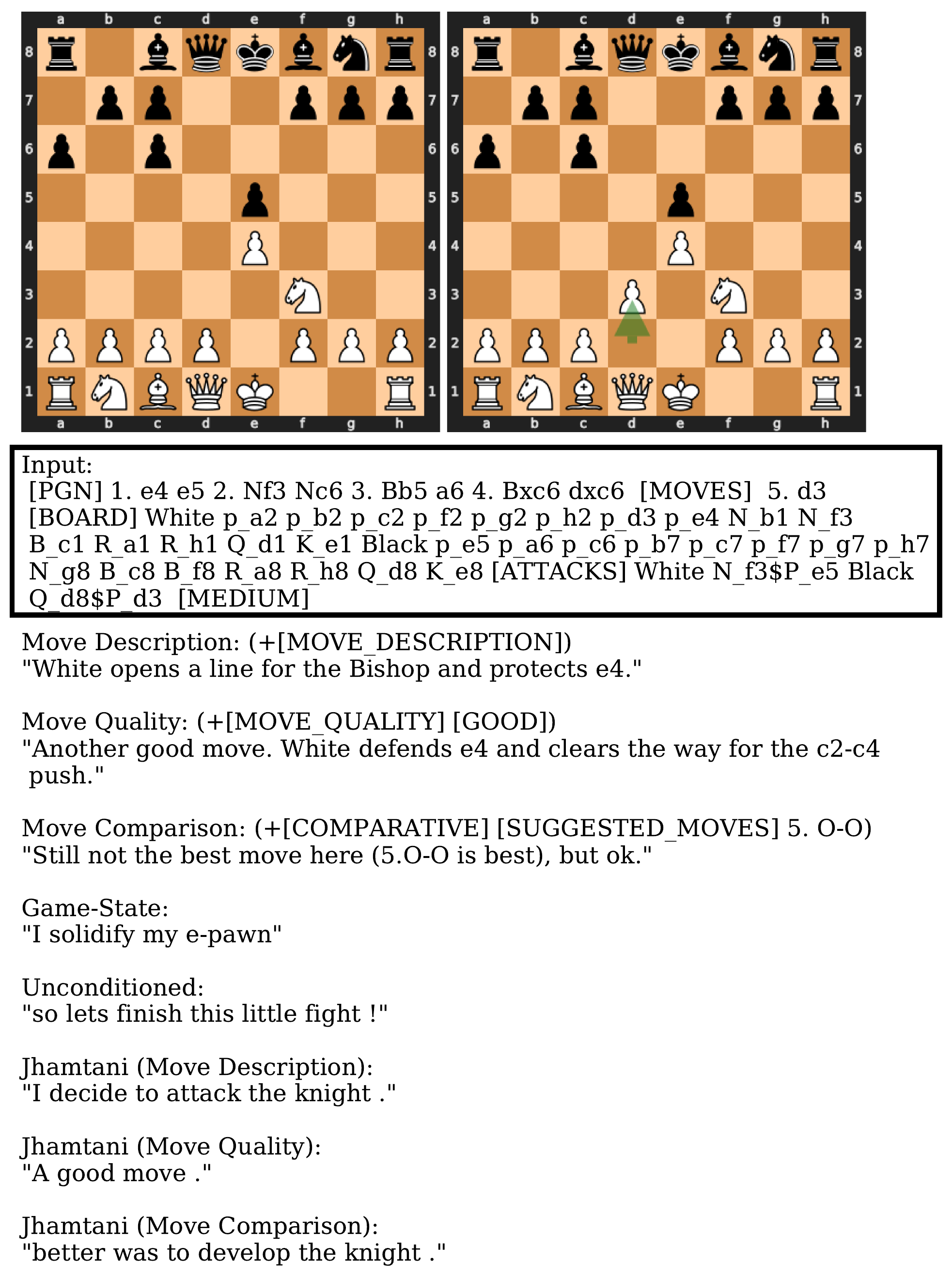}
  \caption{\label{appx_fig:response_samples_7}
All 3 responses from our model are accurate in this example.
}
\end{figure*}

\begin{figure*}[t]
  \includegraphics[clip, trim=0cm 0cm 0cm 0cm, width=0.95\textwidth]{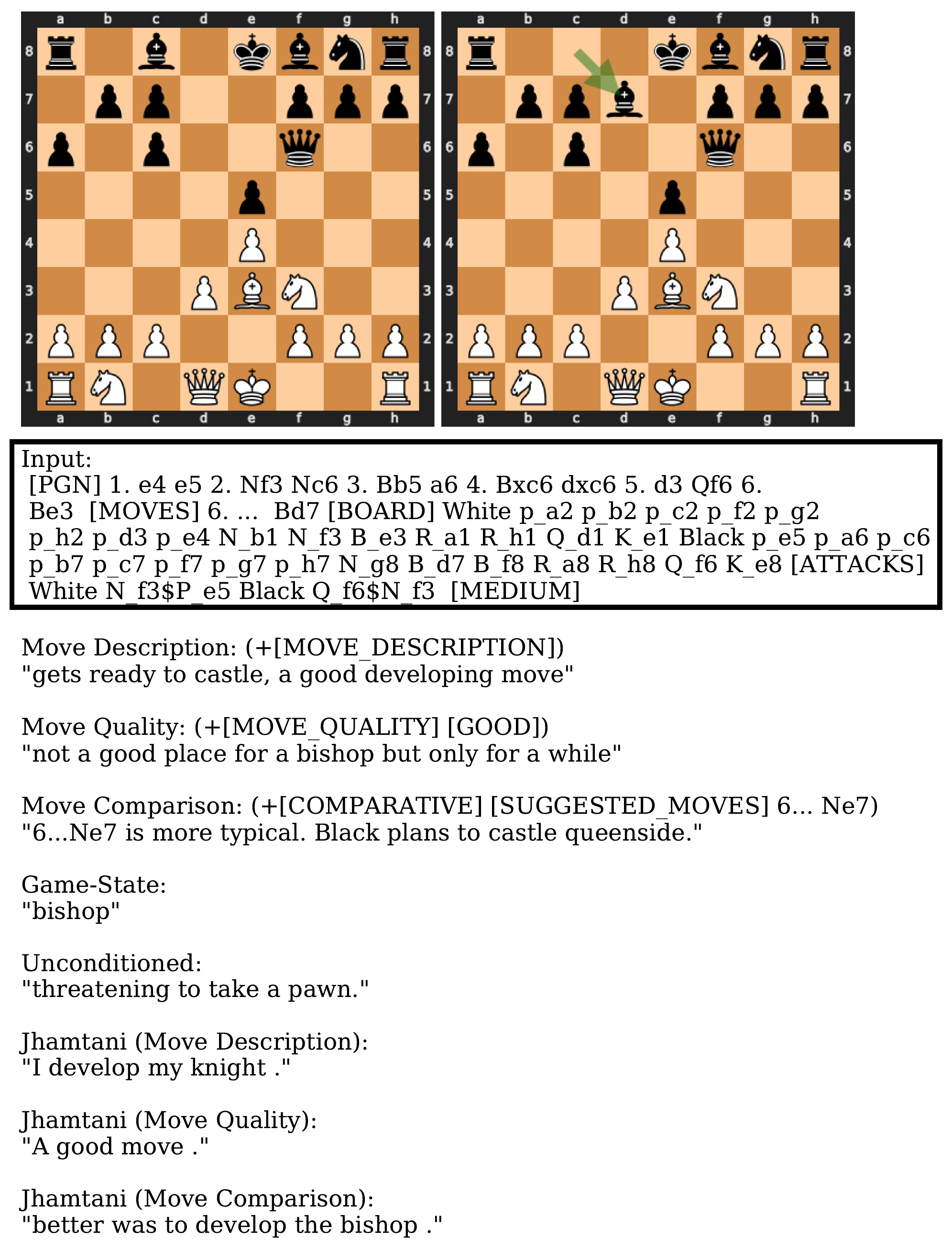}
  \caption{\label{appx_fig:response_samples_8}
All 3 responses from our model are accurate in this example.
}
\end{figure*}

\begin{figure*}[t]
  \includegraphics[clip, trim=0cm 0cm 0cm 0cm, width=0.95\textwidth]{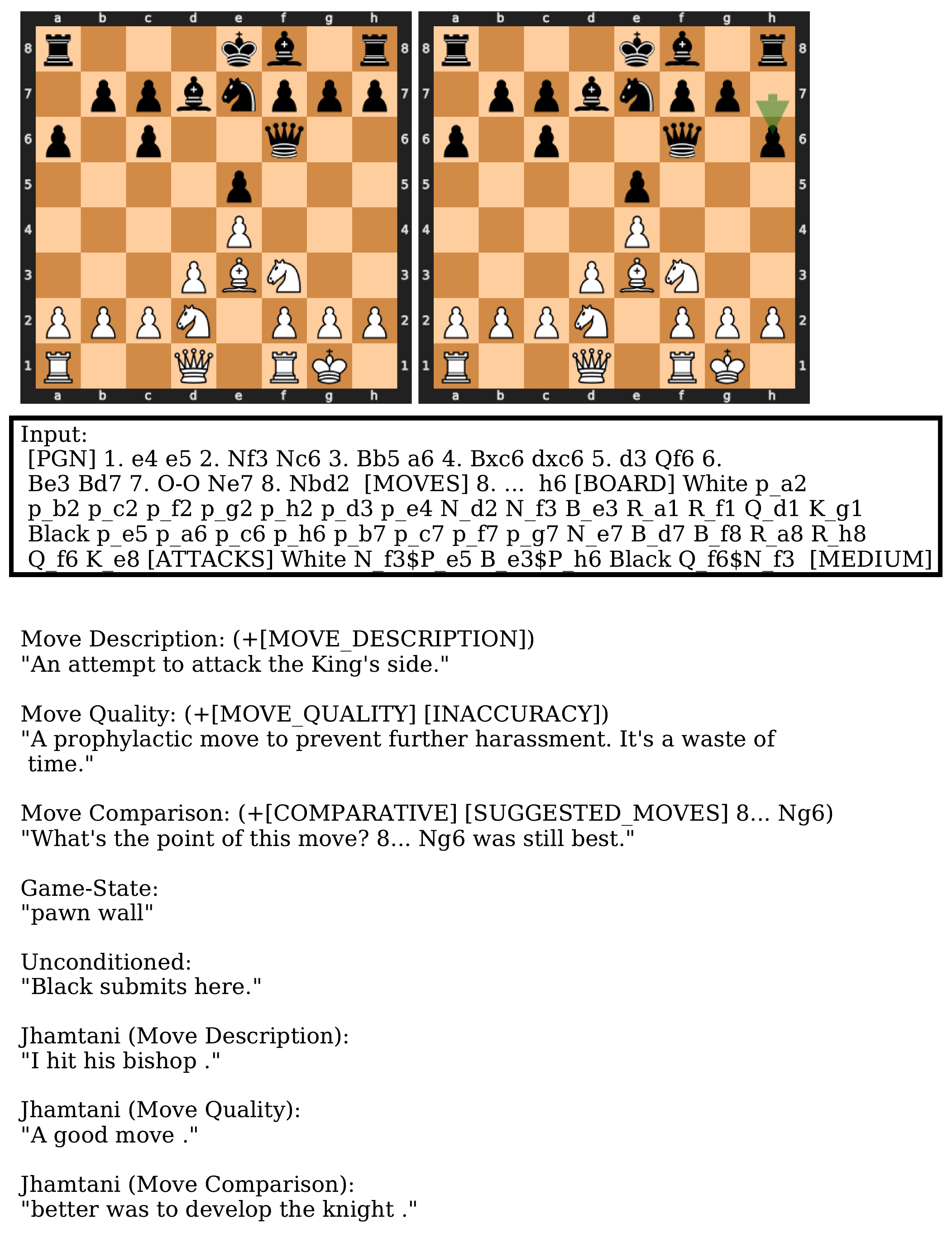}
  \caption{\label{appx_fig:response_samples_9}
All 3 responses from our model are accurate in this example.
Note that for Move Quality, the model describes the move as prophylactic -- likely because the pawn prevents the bishop from moving to g5 to attack Black's queen.
}
\end{figure*}

\begin{figure*}[t]
  \includegraphics[clip, trim=0cm 0cm 0cm 0cm, width=0.95\textwidth]{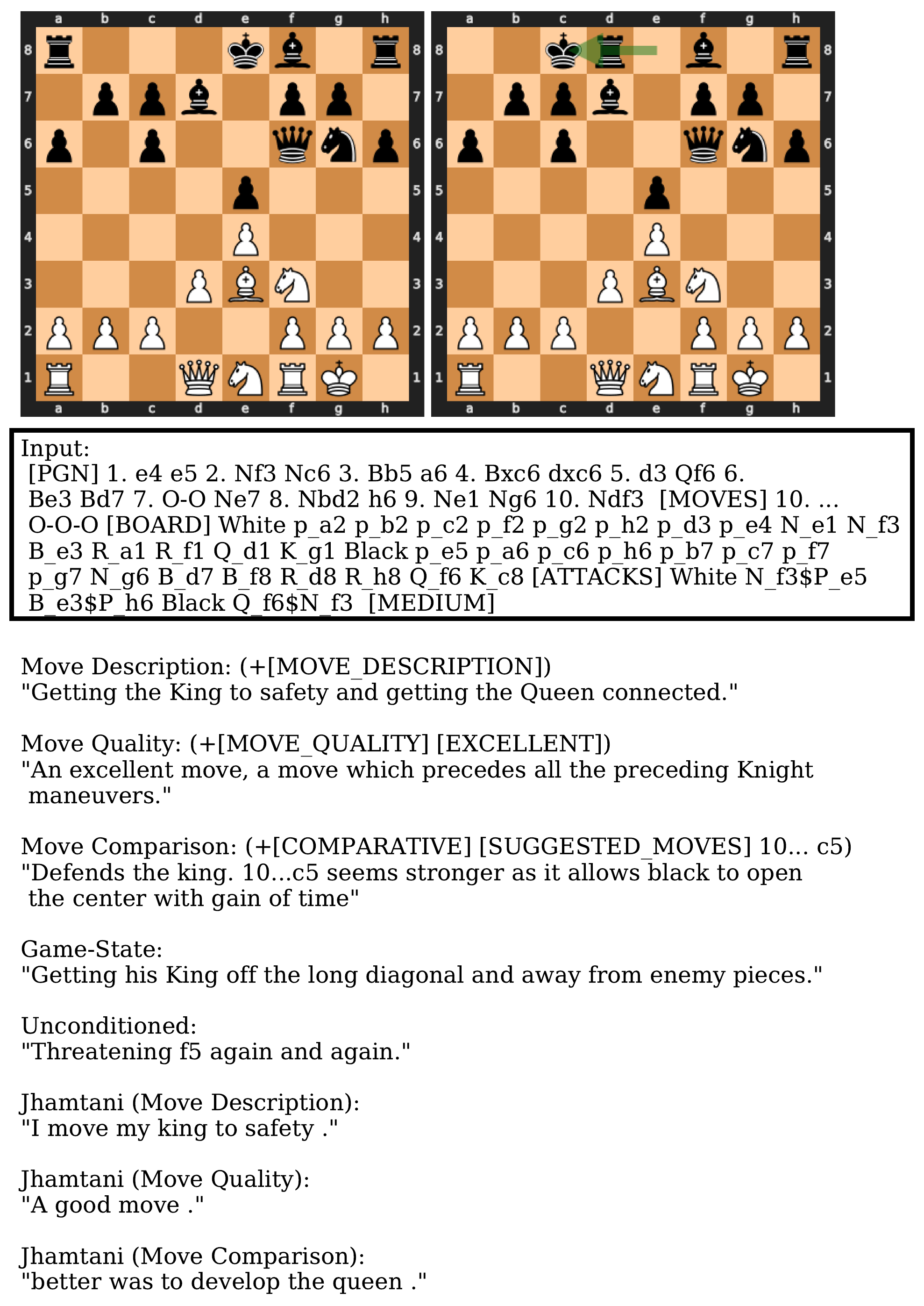}
  \caption{\label{appx_fig:response_samples_10}
For Move Description, our model correctly mentions moving the king to safety, but incorrectly identifies having the queen connected.
This might be a spurious pattern that the model has learned, given that when Black castles and its queen is usually on the 8th rank during the first few moves, the model may think that castling leads to connecting the queen.
The other commentaries are accurate.
}
\end{figure*}

\begin{figure*}[t]
  \includegraphics[clip, trim=0cm 0cm 0cm 0cm, width=0.95\textwidth]{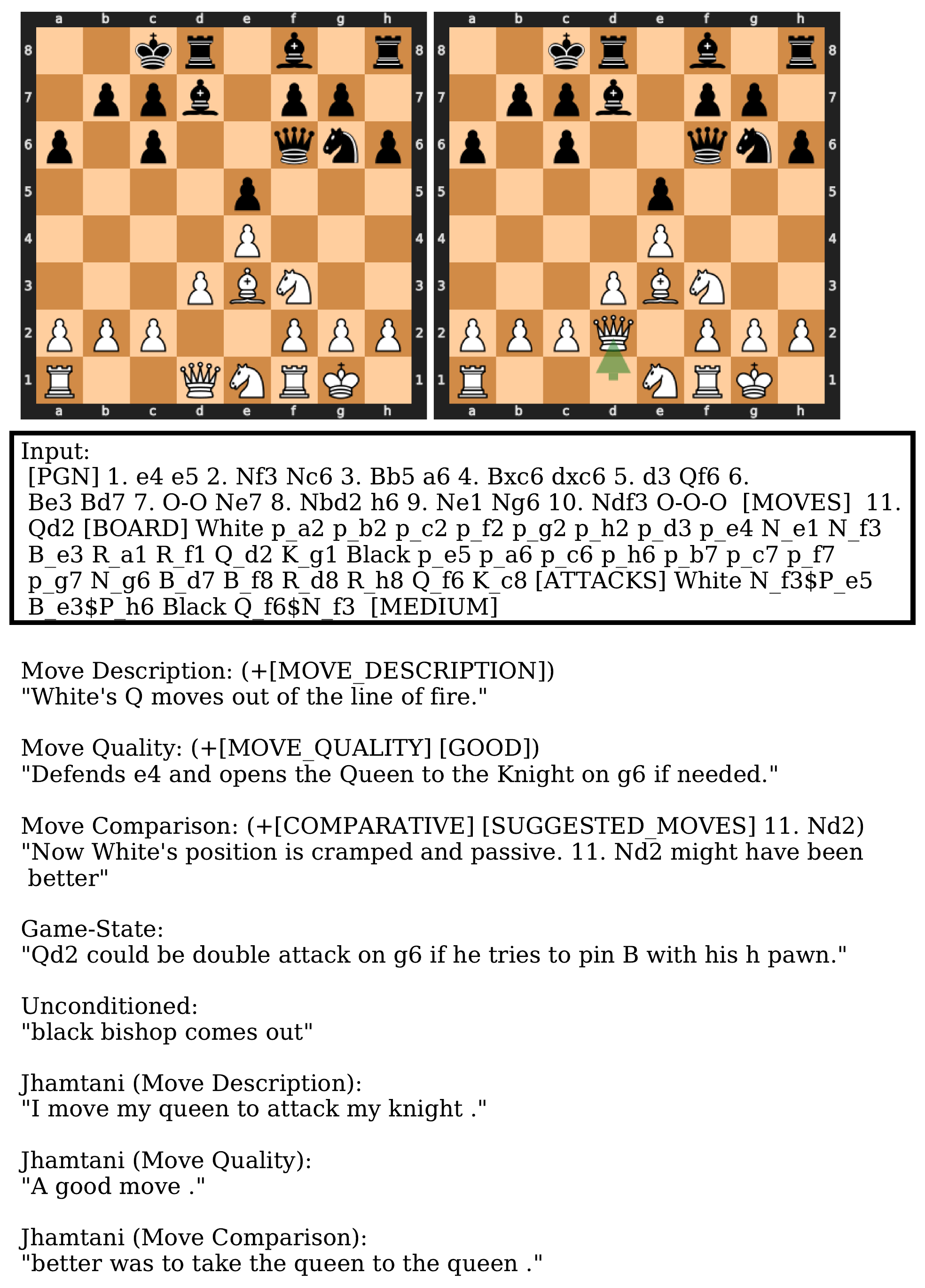}
  \caption{\label{appx_fig:response_samples_11}
For Move Description, the model incorrectly commentates about moving the queen out of the line of fire.
This may be a spurious pattern that the model has learned, as this queen movement is often made if it had been pinned (i.e., if black's bishop had been on g4).
For Move Quality, the model makes multiple logical errors.
}
\end{figure*}

Figure~\ref{appx_fig:response_samples_0} to Figure~\ref{appx_fig:response_samples_11} include samples of responses by our models, with additional comments from the authors regarding the behavior of each model.
For each figure, the first 3 responses are from our fully conditioned model, in which we additionally annotate the tags from Leela that were used to control the generation.

\section{Prompted Belief-States}
\label{appx:heatmaps}
\begin{table}
\small
\centering
\begin{tabular}{l|cc}
    \toprule
       Input Representation & \makecell{Weight\\Distribution} & Accuracy  \\
       
    \midrule
    Random                      & 0.016 & 0.016 \\
    Unconditioned               & 0.082 & 0.154 \\
    Game-State (PGN)            & 0.313 & 0.428  \\
    Game-State (Board)          & 0.657 & 0.806 \\
    Game-State (PGN + Board)    & 0.658 & 0.791 \\
    Fully Conditioned           & 0.589 & 0.726 \\
    \bottomrule
\end{tabular}
\caption{\label{tab:heatmap_metrics}
Accuracy of each model's prompted belief-state.
While a simple board representation demonstrates the most accurate belief-state, it is unable to provide any deeper analyses, such as move suggestions.
}
\end{table} 

Here we provide more details regarding the prompted belief state experiments described in \autoref{sec:discuss}.
We use prompts to derive our model's belief-state -- for each piece on the board, we use a prompt in which we expect a square on the board to follow (``White's bishops on (e5)''), and compute the likelihood of all 64 chess squares to follow.
That is, given model $\mu$, prompt $\pi$, and game-sate $G$, for all 64 chess squares $s \in S$, we compute $P(s|G, \pi; \mu)$, which is then normalized to sum to 1.

Given a belief-state for prompt $\pi$, we compute 1) the amount of weight accumulated on valid squares for $\pi$, and 2) assuming our model follows $\pi$ with token $s' = argmax(P(s): s \in S)$, the percentage of times $s'$ is correct with respect to the game-state.
We use up to 12 prompts per game-state (2 players * 6 chess pieces), excluding cases in which the prompted piece is not on the board.

Table~\ref{tab:heatmap_metrics} shows the results when tested on 500 random game-states not seen during training.
Although our fully conditioned model generates better commentaries, using a simple board representation demonstrates the best belief-state. Given that our input representations are ``additive'', it is curious that additional input signals actually worsens the model's belief-state, hinting that some input tokens are being misused.

Visualizing prompted belief-states may shed some light.
For instance, Figure~\ref{fig:heatmaps} suggests that recent moves impact the model's belief-state: rather than a uniform distribution of weight across all valid squares, most of the weight is concentrated around recently moved pieces.
Additionally, the model occasionally identifies the opponent's pieces when they recently moved (Figure~\ref{fig:heatmaps} (b)).

Intuitively, this behavior makes sense -- commentaries likely discuss recently played moves.
Broadly speaking, in a symbolic, closed domain like chess, we believe prompted belief-states can provide insights for the behavior of language models.

\begin{figure*}[t]
  \centering
  \includegraphics[clip, width=0.98\textwidth]{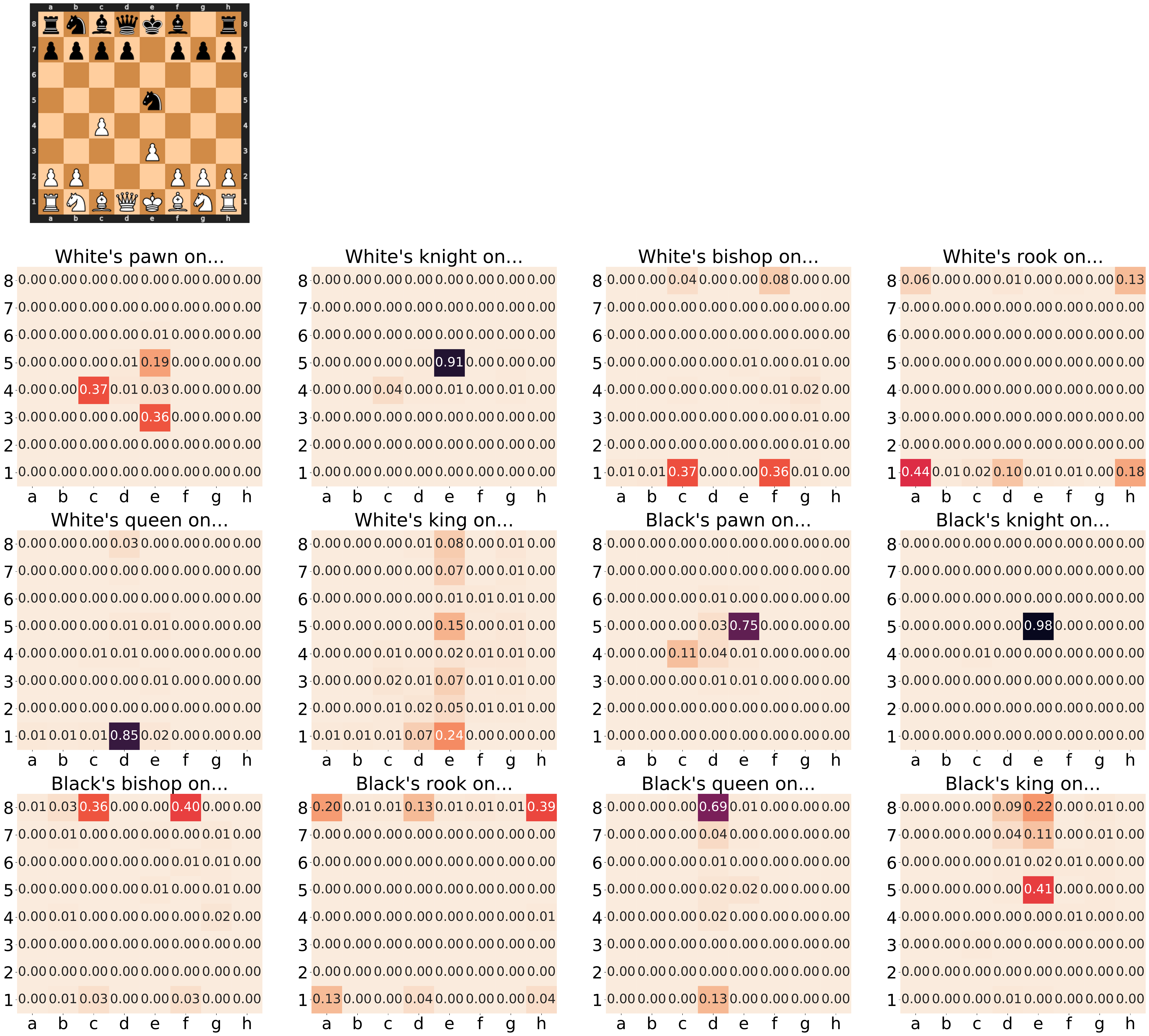}
  \caption{\label{appx_fig:heatmap_0}
Additional prompted belief-states of our model.
}
\end{figure*}

\begin{figure*}[t]
  \centering
  \includegraphics[clip, width=0.98\textwidth]{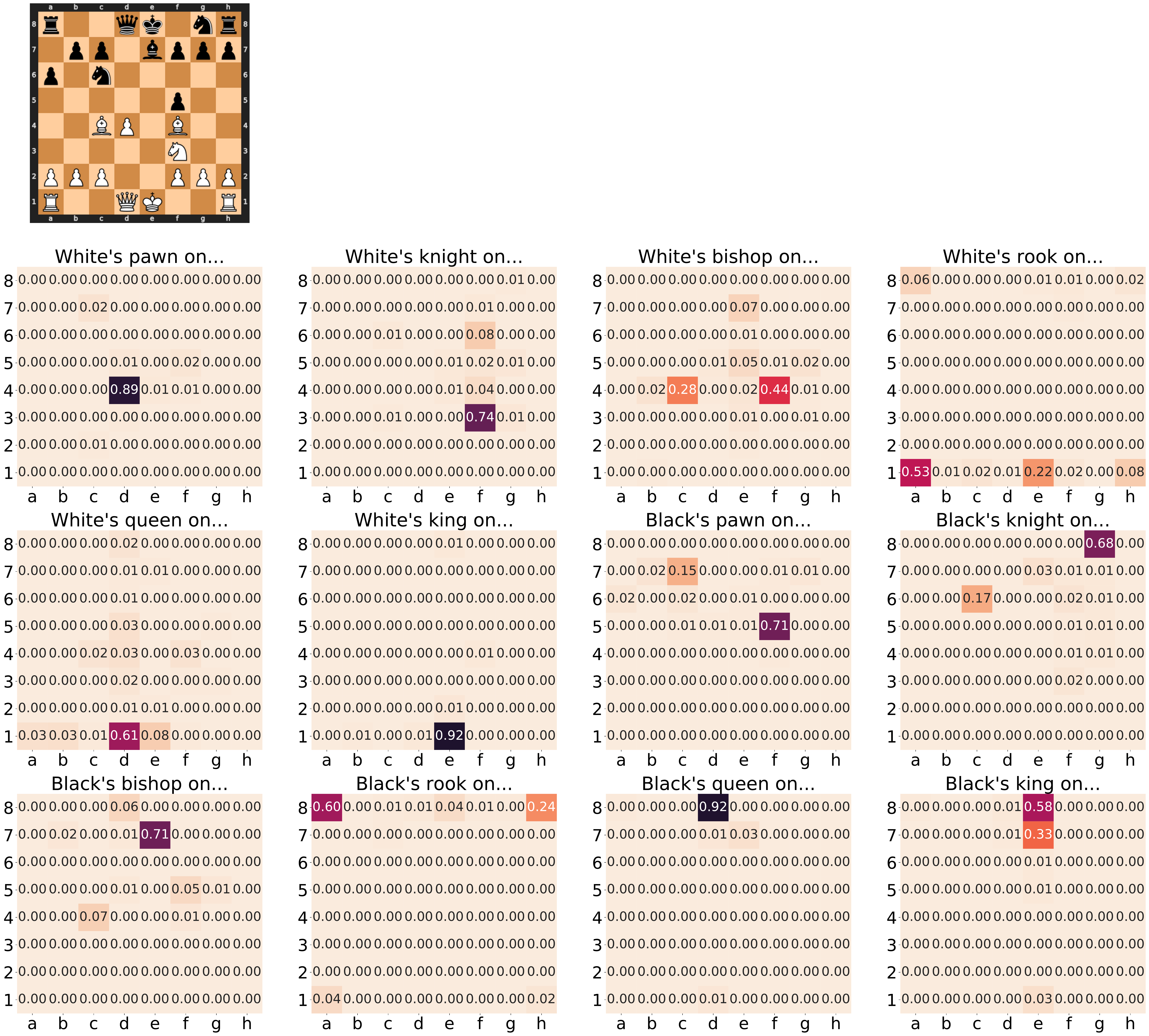}
  \caption{\label{appx_fig:heatmap_1}
Additional prompted belief-states of our model.
}
\end{figure*}

\begin{figure*}[t]
  \centering
  \includegraphics[clip, width=0.98\textwidth]{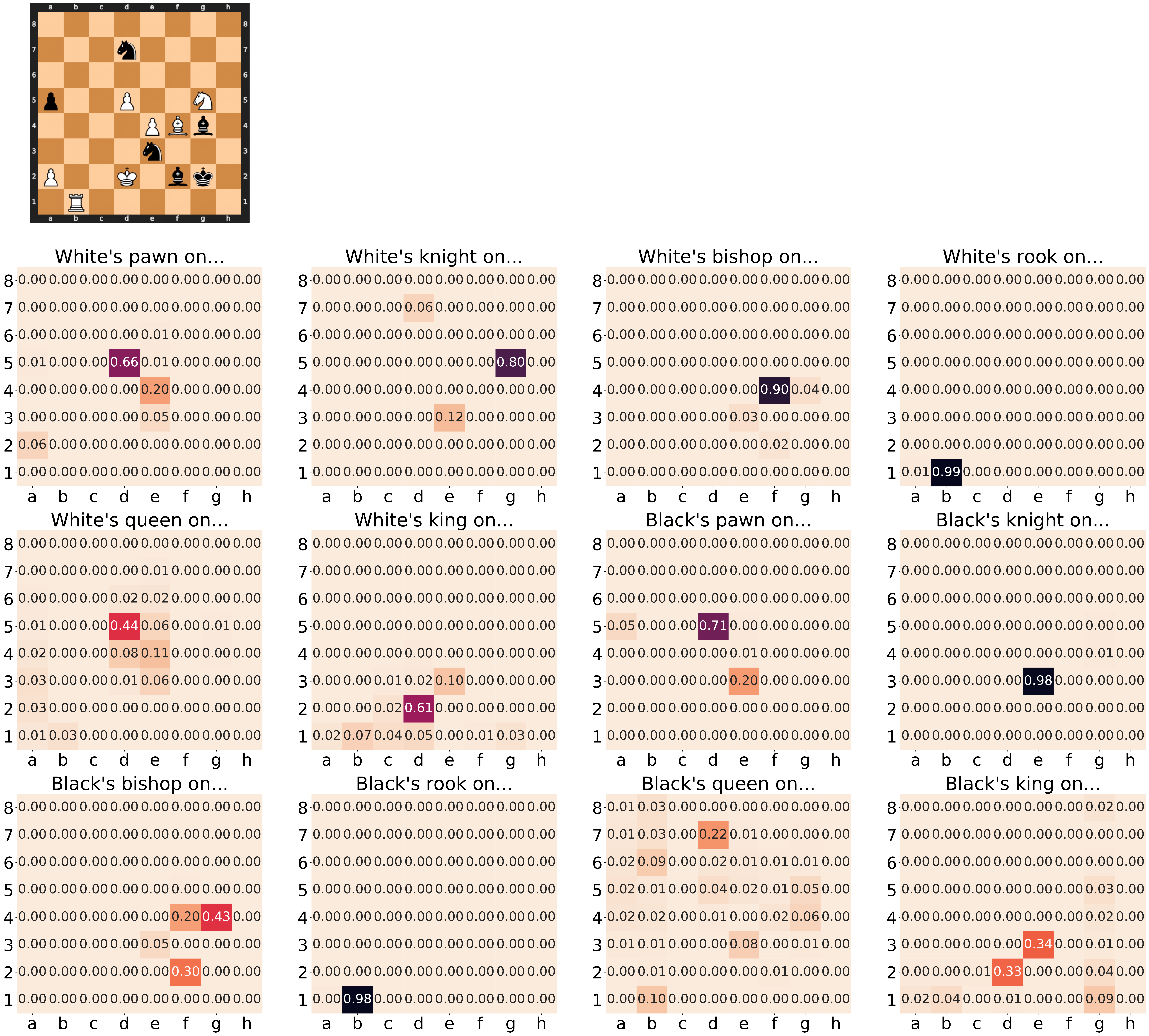}
  \caption{\label{appx_fig:heatmap_2}
Additional prompted belief-states of our model.
}
\end{figure*}

\begin{figure*}[t]
  \centering
  \includegraphics[clip, width=0.98\textwidth]{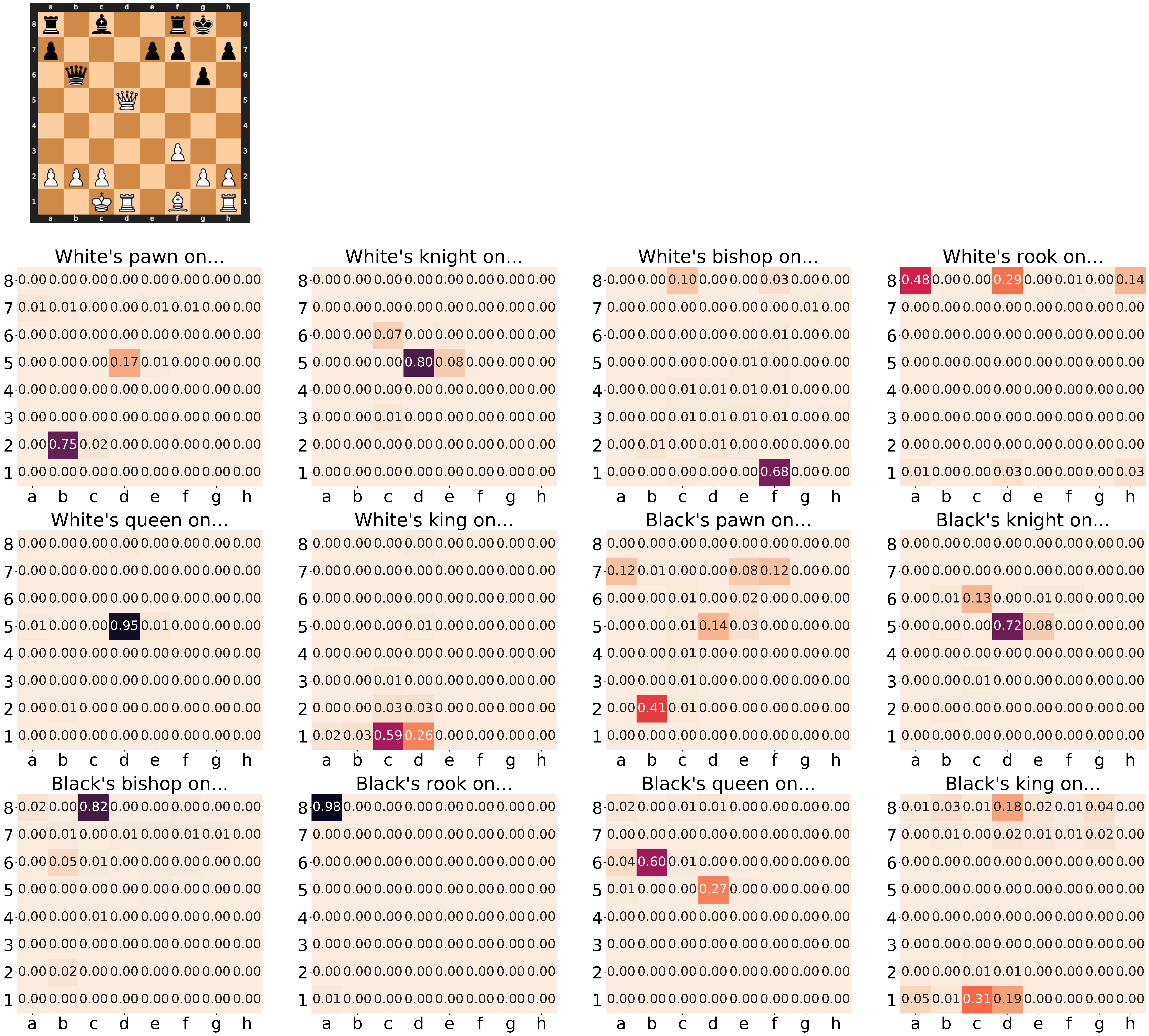}
  \caption{\label{appx_fig:heatmap_3}
Additional prompted belief-states of our model.
}
\end{figure*}

Figure~\ref{appx_fig:heatmap_0} to Figure~\ref{appx_fig:heatmap_3} include additional examples of prompted belief-states of our models.

\begin{table}
\small
\centering
\renewcommand{\tabcolsep}{3pt}
\begin{tabular}{ll}
    \toprule
    Activation & gelu \\
    Attention Dropout & 0 \\
    Batch Size & 2, \textbf{4}, 8, 16 \\
    Dropout & 0.1 \\
    Embedding Size & 1024 \\
    Gradient Clip & 0.1 \\
    Num. Decoder Layers & 12 \\
    Num. Encoder Layers & 12 \\
    Num. Attention Heads & 16 \\
    Num. Transformer Layers & 2 \\
    Optimizer & Adam \\
    Adam Eps & 1e-8 \\
    Learning Rate Scheduler & Linear \\
    Learning Rate Scheduler Decay & 0.5 \\
    Learning Rate Scheduler Patience & 20 \\
    Learning Rate & \textbf{1e-4}, \textbf{1e-5}, 1e-6 \\
    Validation Metric & Loss \\
    Warmup Rate & 1e-4 \\
    Warmup Updates & 3000 \\
    \bottomrule
\end{tabular}
\caption{\label{tab:hyperparams_pretrain}
Hyperparameters for our pre-trained BART model.
}
\end{table} 

\begin{table}
\small
\centering
\renewcommand{\tabcolsep}{3pt}
\begin{tabular}{ll}
    \toprule
    Activation & gelu \\
    Attention Dropout & 0 \\
    Batch Size & \textbf{4}, 8, 16, 32 \\
    Dropout & 0.1 \\
    Embedding Size & 1024 \\
    Gradient Clip & 0.1 \\
    Num. Decoder Layers & 12 \\
    Num. Encoder Layers & 12 \\
    Num. Attention Heads & 16 \\
    Num. Transformer Layers & 2 \\
    Optimizer & Adam \\
    Adam Eps & 1e-8 \\
    Learning Rate Scheduler & Linear \\
    Learning Rate Scheduler Decay & 0.5 \\
    Learning Rate Scheduler Patience & 20 \\
    Learning Rate & 1e-4, \textbf{1e-5}, 1e-6 \\
    Validation Metric & Loss \\
    Warmup Rate & 1e-4 \\
    Warmup Updates & 3000 \\
    \bottomrule
\end{tabular}
\caption{\label{tab:hyperparams_comm_type}
Hyperparameters for commentary-type classification model.
}
\end{table} 

\begin{table}
\small
\centering
\renewcommand{\tabcolsep}{3pt}
\begin{tabular}{ll}
    \toprule
    Activation & gelu \\
    Attention Dropout & 0 \\
    Batch Size & 4, 8, 16, \textbf{32} \\
    Dropout & 0.1 \\
    Embedding Size & 1024 \\
    Gradient Clip & 0.1 \\
    Num. Decoder Layers & 12 \\
    Num. Encoder Layers & 12 \\
    Num. Attention Heads & 16 \\
    Num. Transformer Layers & 2 \\
    Optimizer & Adam \\
    Adam Eps & 1e-8 \\
    Learning Rate Scheduler & Linear \\
    Learning Rate Scheduler Decay & 0.5 \\
    Learning Rate Scheduler Patience & 20 \\
    Learning Rate & \textbf{1e-4}, 1e-5, 1e-6 \\
    Validation Metric & Loss \\
    Warmup Rate & 1e-4 \\
    Warmup Updates & 3000 \\
    \bottomrule
\end{tabular}
\caption{\label{tab:hyperparams_move_qual}
Hyperparameters for move quality classification model.
}
\end{table} 
\begin{table}
\small
\centering
\renewcommand{\tabcolsep}{3pt}
\begin{tabular}{ll}
    \toprule
    Activation & gelu \\
    Attention Dropout & 0 \\
    Batch Size & \textbf{2}, 4, 8 \\
    Dropout & 0.1 \\
    Embedding Size & 1024 \\
    Gradient Clip & 0.1 \\
    Num. Decoder Layers & 12 \\
    Num. Encoder Layers & 12 \\
    Num. Attention Heads & 16 \\
    Num. Transformer Layers & 2 \\
    Optimizer & Adam \\
    Adam Eps & 1e-8 \\
    Learning Rate Scheduler & Linear \\
    Learning Rate Scheduler Decay & 0.5 \\
    Learning Rate Scheduler Patience & 20 \\
    Learning Rate & 1e-4, 1e-5, \textbf{1e-6}, 1e-7 \\
    Validation Metric & Loss \\
    Warmup Rate & 1e-4 \\
    Warmup Updates & 3000 \\
    \bottomrule
\end{tabular}
\caption{\label{tab:hyperparams_sugg_move}
Hyperparameters for suggested moves extraction model.
}
\end{table} 

\begin{table}
\small
\centering
\renewcommand{\tabcolsep}{3pt}
\begin{tabular}{ll}
    \toprule
    Activation & gelu \\
    Attention Dropout & 0 \\
    Batch Size & 2, \textbf{4}, 8 \\
    Dropout & 0.1 \\
    Embedding Size & 1024 \\
    Gradient Clip & 0.1 \\
    Num. Decoder Layers & 12 \\
    Num. Encoder Layers & 12 \\
    Num. Attention Heads & 16 \\
    Num. Transformer Layers & 2 \\
    Optimizer & Adam \\
    Adam Eps & 1e-8 \\
    Learning Rate Scheduler & Linear \\
    Learning Rate Scheduler Decay & 0.5 \\
    Learning Rate Scheduler Patience & 20 \\
    Learning Rate & 1e-4, \textbf{1e-5}, 1e-6 \\
    Validation Metric & PPL \\
    Warmup Rate & 1e-4 \\
    Warmup Updates & 3000 \\
    \bottomrule
\end{tabular}
\caption{\label{tab:hyperparams_comm_gen}
Hyperparameters for commentary generation model.
}
\end{table} 

\begin{table*}
\small
\centering
\renewcommand{\tabcolsep}{3pt}
\begin{tabular}{p{0.97\linewidth}}
    \toprule
    \textbf{Accuracy} (36\%) \\
    \midrule
     "The statements A is making do not make sense. In the first move, A states that \"I have hit his knight\", but the move does not take a knight off the board. For the third move, A states that \"I move my queen to attack my knight\", but one does not attack their own pieces. "\\
    "A is more accurate." \\
    "B is only accurate or relevant in one of the scenarios. " \\
    "B comment at board 2 mentions a black knight move at 18, when the knight was taken earlier." \\
    "A seems to be more accurate in assessing the board." \\
    "A is more precise, while B references incorrect moves and is nonsensical. " \\
    "Agent B describes correctly" \\
    "Agent A describes correctly" \\
    "A is accurate." \\
    "A has more accurate play. B references incorrect pieces. " \\
    "A is slightly better.  B isn't making any sense and/or is ambiguous, although A isn't making much sense, either (i.e. one develops pieces, not pawns)." \\
    "B doesn't make sense. There's a pawn at g4 so the knight couldn't have moved there in set 1. In set 3 it's a broken sentence about bishops that aren't on the board." \\
    "Board 1, queen swap is not forced. Queen could run. I don't understand how A is logical, so =b. Board 2, =b. Board 3, =b. Overall B." \\
    "B mentions inaccurate pieces and lines. " \\
    "A mentions a bishop not on the board. Neither are very accurate, though. " \\
    "Agent B describes correctly" \\
    "A is wordy but not keeping with the current board pieces or positions." \\
    "A has errors, but B does too, and mentions the wrong pieces in \#2" \\
    "A has highlighted more relevant information about the moves" \\
    "A was the most knowledgeable and professional sounding. B was a little flippant and mean at times. " \\
    "Agent B describes correctly" \\
    "Agent B incorrectly mentions that Black has the advantage of two pieces, when they're in fact down by one." \\
    "A seems to be getting their pieces mixed up. B makes sense with what happened on the chess board." \\
    "This comment by A here 'A: Hoping he'll take the queen so I can queen.' is wrong as there are no Queens in the game!. Also wrong 'A: White knows he has an advantage, but has to defend. This move forces the black rook to a square where it can be attacked.' as the knight can be taken down by the king. B is making slightly more sense. " \\
    "B's comments on the 2nd and 3rd moves are nonsense, referencing pieces not moved. " \\
    "Agent B isn't talking about the correct moves at times." \\
    "I don't love either one, but Agent B's responses seem to make more sense." \\
    "Agent A incorrectly mentions White attack pieces which haven't occurred, while B talks about connected pawns by White." \\
    "B mentions illegal lines and incorrect tactics. " \\
    "I don't think A's comments were accurate in 2nd \& 3rd turn. " \\
    "B is make better chess commentaries, it moves correctly" \\
    "A's commentary talks about pieces that are no longer on the board, which I think is a bigger problem that B's problems as commentator." \\
    "Agent A correctly addresses black's retreat and taking of pawn, while Agent B mentioned the attack of its own piece." \\
    "I feel like B had a decent reason in the first move but it was sort of rambling and not too clear while A may be a bit too short/not the most accurate but mostly on point." \\
    "For board 1, A's comment is correct in describing what is happening. B's comment is vague and not very helpful because it lacks detail. For board 2, both comments are okay, but A does a better job by being more detailed and stating what actually happened. For board 3, both comments seem okay in that they are capturing commentary related to the actual move. " \\
    "B appears confused about black's inventory of pieces." \\
\bottomrule
\end{tabular}
\caption{\label{tab:reasons_ours_i}
Reasons that crowdworkers prefer our model (Part I).
}
\end{table*} 

\begin{table*}
\small
\centering
\renewcommand{\tabcolsep}{3pt}
\begin{tabular}{p{0.97\linewidth}}
    \toprule
    \textbf{Strategy} (14\%) \\
    \midrule
        "A gives the logic behind the commentary and how to utilize it." \\
        "not always the best move but the best in these situations, solid moves" \\
        "A gives specific suggested moves.  B is more general with the analysis of the specific move.  " \\
        "B references more correct strategies. " \\
        "While B is too wordy and the suggested move in set 3 doesn't accomplish much, it at least makes sense. A is just repeating 'develop the queen' which is the worse option of the two." \\
        "B more accurately describes the tactics. " \\
        "For board 1, B has a much better comment becuase it is very correct in that white could just move the g pawn up to g2 and the queen would be forced to move. A dosen't seem to make sense in the context because the king has not yet moved off its starting square. For board 2, I am not sure that A is talking about this turn, as that would be a better comment when black is taking a turn given it could move a knight to the position referenced. B's comment picks up on white's queen moving, which is good. For board 3, both comments fit for the most part. I like A better because it is written out in a manner that is easier to understand. B does a better job across all three boards however.", \\
        "Agent B correctly determines analysis of castling, simple developing moves, and overall regard to game-play." \\
        "For board 1, A makes a valid alternative suggestion for the bishop's movement, but I think we are on move 14, not the 13 it srote.. B's comment fits as well. I prefer alternative possibilities as seen with A's comment as I think they are more useful, but both are okay. For board 2, A makes a good comment in that white's knight could be exchanged for black's bishop. B's comment makes a reference to white castling,, but that is not a possibility at this stage of the game. That mistake makes the rest of their suggestion not relevent. For board 3, neither has a good comment as all they do is list a bunch of moves without explaining anything." \\
        "Agent B mentions the end goal of pushing pawn forward to be upgraded." \\
        "Agent B mentions moves being used as a defense and playable while Agent A describes pieces that haven't been recently moved." \\
        "Board 1: Neither. Commentary A doesn't make sense for this board. There are no queens, plus its not white turn. For commentary B, Ne6 is not a better move. It puts it risk and is a sacrifice that does not need to be made.Board 2: B. How is commentary A logical if it's black's move? Board 3: A Overall: B" \\
        "Assistant A is much more thorough in their descriptions about the moves. It describes why a play is good or bad, not just that it is good or bad." \\
        "Both sort of wrong, at least B provided more insight than A." \\
    \midrule
    \textbf{Diversity, Detailed, Other} (8\%) \\
    \midrule
        "A is correctly describing what players are doing on the board with their moves. It is more descriptive. The commentary would be helpful for readers to study the game. B on the other hand isn't very helpful with short statements. " \\
        "A seems to be writing a bit in more detail. However in the last one didn't need to write 'White does not have to be careful'. That didn't make sense. Overall I think both the commentaries could use more elaboration. " \\
        "A is more descriptive, especially in the second and third scenarios.  B indicates a good move every time, but gives no explanation why this is the case." \\
        "Agent B repeat the same comment for all moves. " \\
        "Both are poor and inaccurate, but A is repetitive in \#1, and vague in \#2 and 3. " \\
        "While it might've been better to develop the pawns, Agent A repeats the same critique each move without clarifying." \\
        "B is more fluent and detailed. " \\
        "Referencing time is not helpful in this context. " \\
    \midrule
    \textbf{Non-informative} (42\%) \\
    \midrule
        (Example) ``Just my opinion'' \\
        (Example) ``Because A is better'' \\
    \bottomrule
\end{tabular}
\caption{\label{tab:reasons_ours_ii}
Reasons that crowdworkers prefer our model (Part II).
}
\end{table*} 

\begin{table*}
\small
\centering
\renewcommand{\tabcolsep}{3pt}
\begin{tabular}{p{0.97\linewidth}}
    \toprule
    \textbf{Accuracy} (15\%) \\
    \midrule
        "Board 1:  B Board 2: B Board 3: A. Unsure how Re7 is possible in commentary B. Overall: A. Though I like B for boards 1 \& 2, A is also acceptable commentaries for those boards. Since board 3 needs to be A, I went A overall." \\
        "Board 1's comments are both fine. For board 2, I like the detail in B's comment more than the simple phrase given by A. Both fit the action however. For board 3, I prefer B's comment because A references a retreat, when technically it advanced. I think the intention of the comment was to say it was getting out of check however, which is indeed what happened. I still like B's better though." \\
        "Board 1, putting knight forward best move=A. Board 2, nonsense. Either choice. Board 3, nonsense. Either choice. I think A is better just for the Board 1." \\
        "I don't feel the last comment was correct by B. They aren't attacking their king and they can't attack the White king by this position.   But the other 2 comments were more accurate than A" \\
        "Both agents suggested at least one move that is either a bad move or doesn't match the board, but I believe A is generally less mistaken and more detailed." \\
        "Board 1: B. Not really attacking, but puts white queen in a more awkward position. Board 2: A Board 3: B Overall: B" \\
        "Agent A seems better to me as a commentator, but not necessarily a knowledgeable one, Agent B comes across as the quieter backup like real sports announcer set-ups." \\
        "Agent A describes correctly" \\
        "Both are quite terrible. A's first 2 comments don't make much sense. And I frankly don't understand B's commentary on the 2nd board. On the 3rd board, A's comment doesn't fit as the only thing the pawn could have done was promote and then get instantly captured by the queen -- however both agents missed the Qc5+ forking the queen and king as the easy winning move" \\
        "Agent A does not gave the correct comment to the move. So B is better" \\
        "First comment by B 'but he didn't take it' doesn't make sense. A talks about 'dice' in the game. Not sure where that is coming from? Also in the third comment A talks about 'Attacker to attacker' could have used better verbiage as in taking the attack to the black etc. Also B makes a better comment in the third iteration Queen should have defended by confronting the other queen. " \\
        "B has better wording, and is more accurate in the last example. " \\
        "Neither Agent is perfect, however Agent B mentions avoiding a loss at a moment when King had multiple choices and plenty of support." \\
        "A did not understand that the pawn was promoted " \\
        "Agent A considers a check that's many moves away and provides only criticism in regards to movement." \\
    \midrule
    \textbf{Strategy} (7\%) \\
    \midrule
        "Both options are bad, but B makes a little more sense, and references a better plan. " \\
        "Agent A seems to understand the game better, and gives a better description of the situation." \\
        "It gives clear and big explanation to the moves." \\
        "Agent A always gave a better advice. But B says exact move. " \\
        "Agent B gave some strong recommendations than A " \\
        "B gives some alternative suggestions and explains the strategy to some degree.  " \\
        "A was much more descriptive with strategy and the overall analysis of the match. " \\
    \midrule
    \textbf{Diversity, Detailed, Fluency, Other} (10\%) \\
    \midrule
        "B is significantly more descriptive in all three scenarios and is easy to understand. " \\
        "Agent B describes moves that aren't deeply complicated and a passive move on the grand scale." \\
        "Agent B provides more specific information and play strategy" \\
        "Agent B describes deeply" \\
        "Neither are very good. A is simple and repetitive, but B is repetitive and incorrect in \#3. " \\
        "Both were a bit too technical for my taste, but agent (A) seems to be less repetitious. " \\
        "Agent A describes simply" \\
        "Neither is a very good annotation but A has detail and natural language. " \\
        "A provided more coherent commentary in two out of three instances, though it was a bit lacking in the third move's commentary." \\
        "A's first comment was unreadable." \\

    \midrule
    \textbf{Non-informative} (68\%) \\
    \midrule
        (Example) ``Just my opinion'' \\
        (Example) ``Because A is better'' \\    
    \bottomrule
\end{tabular}
\caption{\label{tab:reasons_baselines}
Reasons that crowdworkers prefer baseline models.
}
\end{table*} 



\end{document}